\documentclass{article}

\usepackage[final]{corl_2019} 

\usepackage{soul} 
\usepackage{graphicx}
\usepackage{subcaption}
\usepackage{amsmath,amssymb}
\usepackage{todonotes}
\usepackage{floatrow}
\usepackage{siunitx}
\usepackage{bbm}
\usepackage{dsfont}
\usepackage[ruled,vlined,linesnumbered]{algorithm2e}
\usepackage{booktabs}
\usepackage{enumitem}
\usepackage[title]{appendix}

\newcommand{\highlevel}{\mathit{HL}}
\newcommand{\lowlevel}{\mathit{LL}}

\newcommand*{\inlineequation}[2][]{%
  \begingroup
    \refstepcounter{equation}%
    \ifx\\#1\\%
    \else
      \label{#1}%
    \fi
    \relpenalty=10000 %
    \binoppenalty=10000 %
    \ensuremath{%
      #2%
    }%
    \@eqnnum
  \endgroup
}

\let\OLDthebibliography\thebibliography
\renewcommand\thebibliography[1]{
  \OLDthebibliography{#1}
  \setlength{\parskip}{0pt}
  \setlength{\itemsep}{2pt plus 0.6ex}
}

\makeatother

\title{HRL4IN: Hierarchical Reinforcement Learning for Interactive Navigation with Mobile Manipulators}

\author{
  Chengshu Li~~~~~Fei Xia~~~~~Roberto Mart\'in-Mart\'in~~~~~Silvio Savarese \\
  Stanford University \\
  \texttt{\{chengshu, feixia, robertom, ssilvio\}@stanford.edu} \\
}

\begin{document}
\maketitle

\vspace{-20pt}


\begin{abstract}
Most common navigation tasks in human environments require auxiliary arm interactions, e.g. opening doors, pressing buttons and pushing obstacles away. This type of navigation tasks, which we call \textit{Interactive Navigation}, requires the use of mobile manipulators: mobile bases with manipulation capabilities. Interactive Navigation tasks are usually long-horizon and composed of heterogeneous phases of pure navigation, pure manipulation, and their combination. Using the wrong part of the embodiment is inefficient and hinders progress. We propose \textbf{HRL4IN}, a novel \textbf{\underline{H}}ierarchical \textbf{\underline{RL}} architecture for \textbf{\underline{I}}nteractive \textbf{\underline{N}}avigation tasks. HRL4IN exploits the exploration benefits of HRL over flat RL for long-horizon tasks thanks to temporally extended commitments towards subgoals. Different from other HRL solutions, HRL4IN handles the heterogeneous nature of the Interactive Navigation task by creating subgoals in different spaces in different phases of the task. Moreover, HRL4IN selects different parts of the embodiment to use for each phase, improving energy efficiency. We evaluate HRL4IN against flat PPO and HAC, a state-of-the-art HRL algorithm, on \textit{Interactive Navigation} in two environments - a 2D grid-world environment and a 3D environment with physics simulation. We show that HRL4IN significantly outperforms its baselines in terms of task performance and energy efficiency. More information is available at \url{https://sites.google.com/view/hrl4in}.
\end{abstract}

\keywords{Hierarchical Reinforcement Learning, Mobile Manipulation, Interactive Navigation}

\vspace{-5pt}
\section{Introduction}
\label{s:intro}
\vspace{-5pt}

In recent years, agents learning with reinforcement have achieved new levels of proficiency, solving from complex games~\cite{silver2017mastering, gu2017deep} and video games~\cite{mnih2013playing, vinyals2017starcraft}, to real-world robotic tasks~\cite{andrychowicz2018learning, kalashnikov2018qt}. In robotics, this success has mostly been restricted to either navigation~\cite{zhu2017target,tai2017virtual, chiang2019learning,faust2018prm} or manipulation with stationary arms~\cite{gu2017deep, vevcerik2017leveraging, levine2018learning, levine2016end, beyret2019dottodot}. The goal in navigation is to change the agent's location without changing the environment's configuration, whereas the goal in manipulation with stationary arms is to change the environment's configuration without changing the agent's location. However, many tasks in real human environments require an agent capable of achieving both, sometimes even simultaneously.

A common case is robot navigation in realistic human environments. This task requires interactions with the environment such as opening doors or pushing obstacles away. We call the family of navigation tasks that require interactions \textit{Interactive Navigation}. Solving interactive navigation problems requires a mobile base with interaction capabilities, i.e. a \textit{mobile manipulator}. With this embodiment, the agent can change its location and/or the configuration of the environment, by using only the base, the arm(s) or a combination of both. A naive use of the entire embodiment for every phase of the task is inefficient: large parts of the task can be solved with only the base or the arm. Therefore, an efficient solution for Interactive Navigation needs to, first, understand what phases of the task are pure navigation, manipulation or both, and second, solve these subtasks efficiently.

The structure mentioned above is well suited for a hierarchical reinforcement learning (HRL) solution, where the high level learns the phase type and sets a subgoal, and the low level learns to achieve it. However, most existing HRL approaches focus on subgoals that are in the same space as the final goal, e.g. intermediate locations towards a final location~\cite{nachum2018data, kulkarni2016hierarchical, sutton1999between} or intermediate joint configurations towards a final one~\cite{levy2018learning}. In our problem setup, however, the subgoals are \textit{heterogeneous}: while the final goal is to move the base to the desired location, the intermediate subgoals may require the agent to move its end-effector to a given pose, e.g., to the location of a door handle. 

In this work, we present \textbf{HRL4IN}, a hierarchical reinforcement learning solution for heterogeneous subtask spaces that can be applied to solve Interactive Navigation tasks. Our solution learns the most efficient use of pure navigation, pure manipulation, or their combination. The key for efficiency is to allow the high level to decide not only on the subgoals for the low level, but also on the capabilities that the low level is allowed to use to achieve it.

In summary, the main contributions of our work are: first, we propose HRL4IN, a novel learning algorithm suited for mobile manipulators. Second, our solution is, to the best of our knowledge, the first deep HRL approach for continuous control that shows success in heterogeneous tasks - tasks with multiple phases where the goals lay on different spaces. Third, we show that HRL4IN can solve Interactive Navigation tasks, navigation tasks that require interaction with the environment. We believe that explicitly defining and addressing this new type of navigation task is relevant and practically impactful for robots that aim to operate in human environments. 

\vspace{-5pt}
\section{Related Work}
\label{s:rw}
\vspace{-5pt}

{\bf Hierarchical Reinforcement Learning:} 
One of the hardest types of tasks for reinforcement learning is long-horizon tasks with sparse reward. Exploring the environment through random sequences of actions requires a prohibitive number of samples until a solution can be found. To alleviate this problem, researchers have proposed hierarchical RL to incorporate temporal abstraction~\cite{sutton1999between}. 

\citet{bacon2017option} proposes an option-critic architecture that learns temporally extended options and their termination conditions, together with the policy over them. \citet{heess2016learning} pretrains the low-level policy on simple tasks and then trains the high-level policy to output appropriate signals that modulate low-level behaviors. Our proposed method, HRL4IN, is closer to the hierarchical approaches where a high-level policy explores by setting subgoals for a low-level policy to achieve~\cite{kulkarni2016hierarchical, vezhnevets2017feudal, nachum2018data}. The subgoals can be set directly in the same space as the original goal~\cite{nachum2018data}, or in an embedding space~\cite{vezhnevets2017feudal}. While the former is more data efficient, the latter has the potential to learn more informative representation. \citet{nachum2018near} propose a way to balance between representation efficiency and capabilities.

Similar to~\cite{nachum2018data, levy2018learning}, our subgoals represent a desired observation that the high-level policy wants to induce. However, we also exploit the fact that a heterogeneous task can be solved more efficiently by setting subgoals in different spaces during different phases of the task. Specifically, our high-level policy not only outputs a subgoal, but also selects a part of the embodiment to use for the current phase. Finally, unlike~\cite{kulkarni2016hierarchical}, we do not manually construct subgoals apriori, which may require extensive domain knowledge, allowing the high-level policy to learn subgoals from scratch.

{\bf Interactive Navigation:}
While the literature on autonomous robot navigation is vast and prolific~\cite{desouza2002vision, bonin2008visual, chenbehavioral}, less attention has been paid to navigation problems that require interactions with the environment, what we call Interactive Navigation. In the robot control literature, several papers have approached the problem of opening doors with mobile manipulators~\cite{peterson2000high, schmid2008opening, petrovskaya2007probabilistic, jain2009behavior}. However, these approaches focus on this single phase and not on the entire Interactive Navigation task.

\citet{stilman2005navigation} studied interactive navigation from a geometric motion planning perspective. In their problem setup, the agent has to reason about the geometry and arrangement of obstacles to decide on a sequence of pushing/pulling actions to rearrange them to allow navigation. They named this type of problems Navigation Among Movable Objects (NAMO)~\cite{stilman2008planning, stilman2007manipulation, van2009path, levihn13}. Their solution requires knowledge of the geometry of the objects to plan, and the search problem is restricted to 2D space.
Our work has two main differences from the previous line of work on NAMO. First, most NAMO solutions assume full observability of the environment as it is required for sampling-based motion planning. On the contrary, HRL4IN only assumes partial observability through virtual depth images. Second, NAMO solutions find a suitable plan in configuration space and rely on a separate controller for the execution of the plan. HRL4IN generates full policies that can not only overcome local minimum in the reward function but also directly output agent commands. The aforementioned properties of our solution render it model-free and end-to-end trainable.

The work by \citet{konidaris2011autonomous} is close to us because it approaches Interactive Navigation as a hierarchical reinforcement learning problem. Their algorithm learns to sequence actions to solve simple tasks and reuses these sequences to solve tasks composed of simple subtasks. However, different from ours, their solution requires a set of predefined controllers as actions for the policy during training. The options are learned in the low-level space and reused during testing. HRL4IN directly controls lower level commands for the mobile manipulator, without hard-coded controllers.

{\bf Task and Motion Planning:}
Interactive Navigation and NAMO can alternatively be solved by task and motion planning (TaMP) methods. Task and motion planning solves long-horizon problems by combining task planner and motion planner. Task planner can reason over large sets of states and motion planner can compute exact paths for the robot to execute. An integration of task and motion planner often uses a hierarchical approach~\cite{wolfe2010combined}. Much of the work in TaMP factorizes the task into sub-problems and abstractions that the solutions learn to use. Existing TaMP approaches are capable of solving complicated mobile manipulation tasks~\cite{srivastava2014combined, garrett2018ffrob}. \citet{konidaris2018skills} learns symbolic representations by interacting with an environment and observing the effects of the actions. \citet{wolfe2010combined} build a system that reuses optimal solutions to state-abstracted sub-problems across the search space. Our work differs from the above as we don't explicitly formulate our problem as TaMP but rather implicitly allow the high-level policy to learn to do the planning and the low-level policy to learn to do the control. 

While most of the previous work on abstraction selection refers to state abstraction~\cite{van2014efficient} (mapping primitive state space $S$ to an abstract state space $h(S)$), our work performs abstraction selection in the action space. The work by \citet{konidaris2008sensorimotor} relates to us as they also learns abstractions for the action space.

\vspace{-5pt}
\section{Background}
\label{s:phrl}
\vspace{-5pt}

We assume that the underlying problem can be modelled as a discrete-time goal-conditioned MDPs $\mathcal{M} = (\mathcal{S}, \mathcal{G}, \mathcal{A}, R, \mathcal{T}, \gamma)$, where $\mathcal{S}$ is the state space,  $\mathcal{G}$ is the goal space, $\mathcal{A}$ is the action space, $R(s, g, a)$ is the reward function, $\mathcal{T}(s'|s, a)$ is the transition dynamics, and $\gamma \in [0, 1)$ is the discount factor. In our work, $\mathcal{G}$ is the space of final goals for the high level and subgoals for the low level. At each time step $t$, an agent observes state $s_{t}$, executes action $a_t$ sampled from policy $\pi(a_t|s_t, g_t)$, receives from the environment a new observation $s_{t+1}$ and a reward $r_t$. In reinforcement learning, the goal of the agent is to maximize expected discounted return $\mathbb{E}_{a\sim\pi}[\sum_{t=0}^{T-1}\gamma^{t} r_{t}]$.
To do so, a family of RL solutions use the policy gradient theorem~\cite{sutton2000policy} to optimize for the policy parameters $\theta$.
In this work, we adopt Proximal Policy Optimization Algorithms (PPO)~\cite{schulman2017proximal}, a stable and widely adopted on-policy model-free policy gradient algorithm as the building block and baseline for HRL4IN.

\vspace{-5pt}
\section{HRL4IN: Hierarchical RL for Interactive Navigation}
\label{s:hrlin}
\vspace{-5pt}

\setlength{\textfloatsep}{10pt}
\begin{figure}
\centerline{\includegraphics[width=0.8\linewidth]{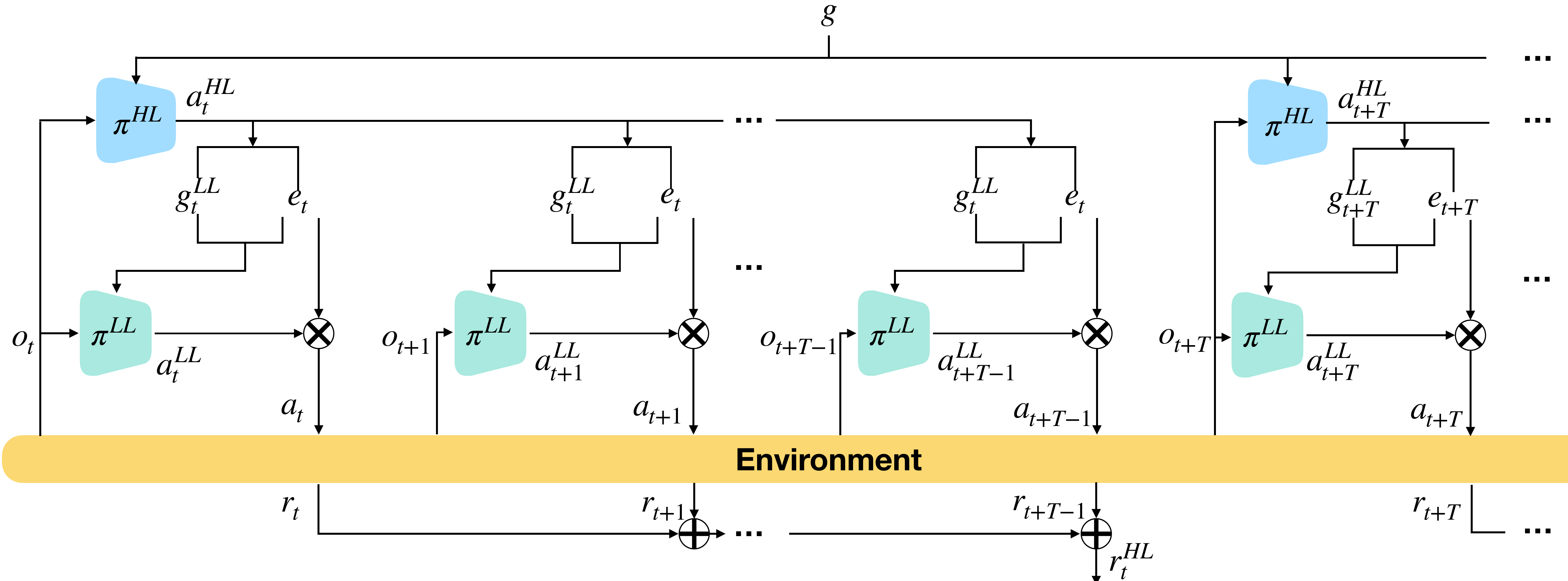}}
\caption{\small{\textbf{HRL4IN}, our Hierarchical Reinforcement Learning solution for Interactive Navigation; both the high-level (blue) and the low-level (green) policies receive an observation from the environment; the high-level policy generates a subgoal and an embodiment selector for the low-level policy for the next $T$ steps, conditioned on the final goal; the low-level policy outputs the robot commands based on the observation, conditioned on the action of the high-level policy; the robot commands are masked and executed by the environment, which gives a reward and the next observation; the sum of these rewards during the $T$ steps will be used to train the high-level policy. More details can be found in Algorithm \ref{alg} in Appendix~\ref{appendix_algo}.}}
\vspace{-10pt}
\label{fig:schema}
\end{figure}

Interactive navigation tasks are intrinsically multi-phase. In some phases the agent only needs to navigate using its base, or to interact with the environment using its arm. In other phases, it needs to move its base and arm in coordination. Using the base or arm when not needed can be counter-productive: it is less energy efficient and increases the probabilities of undesired collisions.

Based on these insights, we propose HRL4IN, a hierarchical reinforcement learning solution for Interactive Navigation that learns to select the elements of the embodiment that are necessary for each phase of the task. To do so, the high-level policy not only sets subgoals for the low-level policy, but also selects which part of the embodiment to use. 

The main structure of HRL4IN is depicted in Fig.~\ref{fig:schema}. Our solution is composed by two policies, a high-level (\textit{HL}) and a low-level (\textit{LL}) policy. Both policies use the same observations $o_{t} \in \mathcal{O}$ that contain both elements that can be changed by the agent (\textit{mutable}) such as the poses of its base and arm, and elements that cannot be changed by the agent (\textit{immutable}) such as the door location. A full description of the observation space is included in Sec.~\ref{ss:setup} and Appendix~\ref{appendix_env}. In the following, we explain the main characteristics of the high-level and the low-level policies.

\vspace{-5pt}
\subsection{High-Level Policy}
\vspace{-5pt}
\label{ss:hlp}


The high-level policy $\pi^{\highlevel}$ acts at a different time scale from that of the original MDP: it acts at time steps $t'$ that correspond to every $T$ time steps of the low-level policy, or fewer, if the low-level policy converges to the subgoal. The high-level policy receives an observation $o_{t'} \in \mathcal{O}$ and generates an action $a^{\highlevel}_{t'} \in \mathcal{A}^{\highlevel}$, conditioned on the final goal, $g \in \mathcal{G}$. 


Similar to previous HRL approaches~\cite{nachum2018data, levy2018learning}, one component of the high-level action, $a^{\highlevel}_{t'}$, is the subgoal $g_{t'}^{\lowlevel} \in \mathcal{G}^{\lowlevel}$. $g_{t'}^{\lowlevel}$ indicates a desired relative change of certain mutable components of the observation, denoted as $x_{t'}$, yielding a subgoal $x_{t'} + g_{t'}^{\lowlevel}$ for the low-level policy to achieve. The subgoal is valid for $T$ time steps of the low-level policy, unless it achieves the subgoal within $T$ time steps ($x_{t}$ is close enough to $x_{t'} + g_{t'}^{\lowlevel}$), in which case HRL4IN queries the high-level policy for the next high-level action. To represent the subgoal with respect to the current observation $x_{t}$, the original subgoal $g_{t'}^{\lowlevel}$ is updated by $g_{t}^{\lowlevel} = x_{t'} + g_{t'}^{\lowlevel} - x_{t}$ for every subsequent time step.

The other component of the high-level action, $a^{\highlevel}_{t'}$, is an embodiment selector $e_{t'}$. The embodiment selector is a discrete variable that decides which capabilities the low-level policy can use to achieve the current subgoal: navigation (base), manipulation (arm), or both. The selector plays two crucial roles. First, it is used to compute an \textit{action mask}, $m^{act}$, that deactivates some components of the embodiment that are not necessary for the current phase (i.e., base or arm) by forcing the velocity of certain joints to be 0. Second, it is used to compute a \textit{subgoal mask}, $m^{sg}$, that determines the dimensions of the observation that the low-level policy should bring closer to the subgoal to obtain \textit{intrinsic reward}. Using the embodiment selector, the high-level policy decides on the phase type within the Interactive Navigation task for a period of $T$ steps: navigation, manipulation or both.

The reward for a high-level action is the sum of the rewards from the environment during the $T$ or fewer than $T$ time steps of its execution. 


\vspace{-5pt}
\subsection{Low-Level Policy}
\vspace{-5pt}

\label{ss:llp}

The low-level policy, $\pi^{\lowlevel}$, acts at every discrete time step $t$ of the MDP. It also receives an observation $o_{t}\in \mathcal{O}$ and generates an action, $a^{\lowlevel}_{t} \in \mathcal{A}^{\lowlevel}$, conditioned on the last high-level action, $a^{\highlevel}_{t'} \in \mathcal{A}^{\highlevel}$, which includes the subgoal $g_{t'}^{\lowlevel}$ (transformed to be relative to the current observation as explained above) and the action and subgoal masks derived from the embodiment selector $e_{t'}$.

The low-level actions, $a^{\lowlevel}_{t}$, are robot commands for the base and the arm. In our Interactive Navigation experiments, the exact form of these actions depends on the environment (see Sec.~\ref{ss:env_setup}). The low-level actions are masked by the action mask derived the embodiment selector $e_{t'}$, which \textit{zeroes out} the action components that are not necessary for the current task phase: $a_{t}=a^{\lowlevel}_{t}\otimes m^{\mathit{act}}_{t'}$. 

We define the current subgoal distance $D_{t}$ as the distance between the current observation $x_{t}$ and the subgoal $x_{t'} + g_{t'}^{\lowlevel}$, masked by the subgoal mask $m^{\mathit{sg}}_{t'}$ derived from the high-level embodiment selector $e_{t'}$: $D_{t} = \sqrt{\sum_{i=0}^{N-1}{m_{t'}^{\mathit{sg}}[i]\cdot(x_{t}[i] - (x_{t'}[i] + g_{t'}^{\lowlevel}[i]))^2}}$, where $N$ is the dimension of $g_{t'}^{\lowlevel}$. We train the low-level policy using intrinsic reward defined by $r^{\lowlevel}_t = D_{t} - D_{t+1}$. Intuitively, the intrinsic reward encourages the low-level policy to make progress towards the subgoal.

For each subgoal, the low-level policy has T time steps to achieve it. We consider a subgoal achieved if $D_{t} = 0$ (discrete state space) or $D_{t} < t_{\mathit{sg}}$ (continuous state space). When the subgoal is achieved or the low-level policy runs out of time, HRL4IN queries the high-level policy for the next $a^{\highlevel}_{t'}$. 
\vspace{-5pt}
\section{Experimental Evaluation}
\label{s:exp}
\vspace{-5pt}

\begin{figure}
\centering
\begin{subfigure}[t]{0.24\textwidth}
\includegraphics[height=3cm]{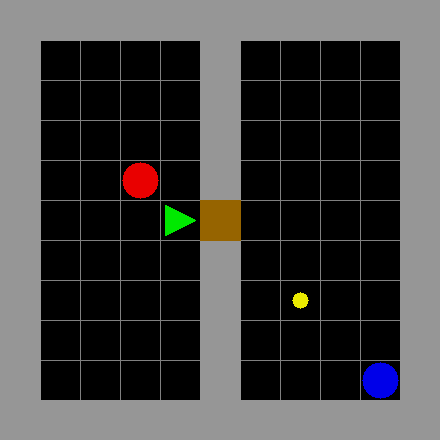}%
\caption{\small{}}
\label{sf:ite}
\end{subfigure}%
\vspace{1pt}
\begin{subfigure}[t]{0.75\textwidth}
\includegraphics[height=3cm]{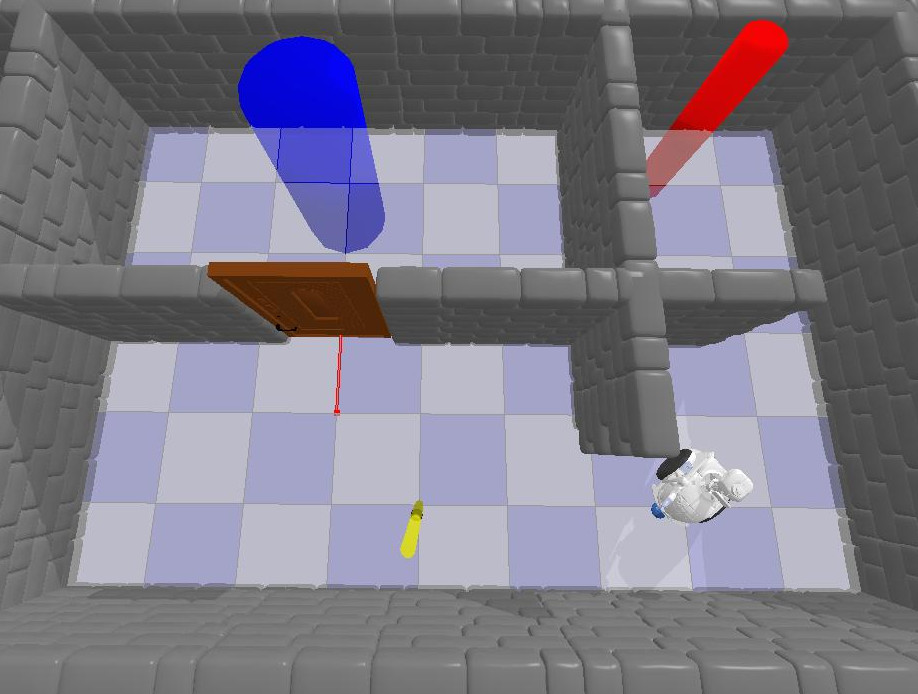}%
\hfill
\includegraphics[height=3cm]{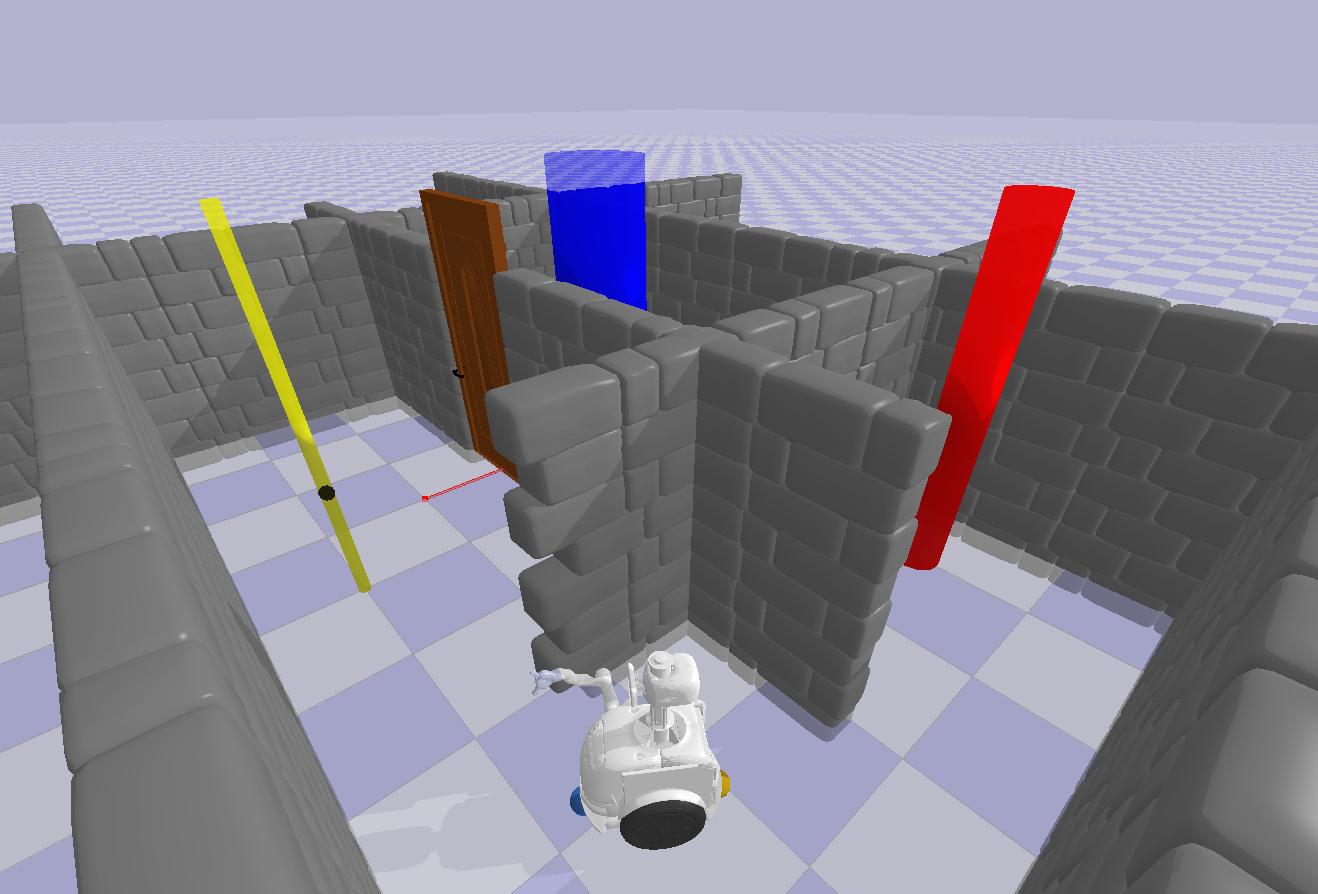}%
\hfill
\includegraphics[height=3cm]{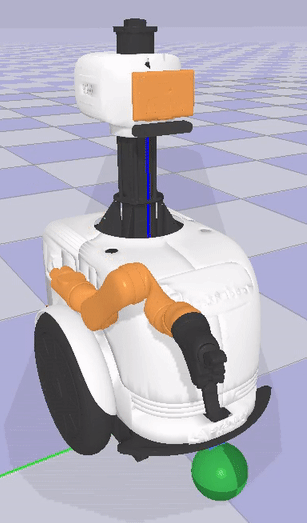}%
\caption{\small{}}%
\end{subfigure}%
\caption{\small{Environments of our experiments; (a) \texttt{Interactive ToyEnv}: the agent (green triangle) has to navigate from the initial location (red circle) to the end location (blue circle) by opening the door that connects the rooms; (b) \texttt{Interactive Gibson Environment}: top-down (left) and perspective (middle) views of the environment where the agent, embodied as a JackRabbot mobile manipulator (right), has to navigate from the initial location (red cylinder) to the end location (blue cylinder) also by opening the door that connects the rooms; In HRL4IN the high-level policy guides the low-level policy by setting subgoals, depicted as a yellow dot in (a) and a yellow cylinder with a black sphere for the desired end-effector position in (b).}}
\label{fig:environments}
\vspace{-10pt}
\end{figure}

\vspace{-5pt}
\subsection{Environment Setup}
\vspace{-5pt}

\label{ss:env_setup}


In order to demonstrate the effectiveness of our approach, we build two environments with different levels of abstraction and physical complexity: \texttt{Interactive ToyEnv} and \texttt{Interactive Gibson Environment}. More details are explained below and included in Appendix~\ref{appendix_env}.


{\bf Interactive ToyEnv:}
\texttt{Interactive ToyEnv} is a 2D grid-world environment with discrete state and action space. It consists of $k\times k$ cells, where each cell could be free space, wall, door, or occupied by the agent. The agent is a simplification of a mobile manipulator that can navigate around the environment and interact with the door using its arm. At each time step, the agent can choose among navigation and/or manipulation actions that result in a deterministic outcome. Navigation actions include \texttt{turn-left}, \texttt{turn-right}, \texttt{go-forward} and \texttt{no-op}. Manipulation actions include \texttt{slide-up} (the door), \texttt{slide-down}, and \texttt{no-op}. Manipulation actions are only effective (i.e., change the door state) when the agent is in the cell in front of the door and is facing the door. Several ($5$) consecutive \texttt{slide-up} are necessary to fully open the door. The observation space includes the global map, the door state, the goal, and proprioceptive information such as the pose of the agent.

{\bf Interactive Gibson Environment:}
\texttt{Interactive Gibson Environment} is a 3D environment with continuous state and action space. It is an evolution of the Gibson Environment~\cite{xia2018gibson} that offers fast rendering and physics simulation. The main difference is that in \texttt{Interactive Gibson Environment}, additional objects (e.g. doors) apart from the agent can be added and their dynamics can be simulated, which allows us to train agents for Interactive Navigation tasks. To simplify the contact physics, we assume the agent has a grasping primitive that grabs the door handle when the end-effector is close enough (distance under $t_{\mathit{grasp}} = \SI{0.1}{\meter}$). Grasping creates a ball joint between the end-effector and the door handle that constrains translation but allows rotation. Once the door is fully opened, the end-effector ends the grasp primitive and releases the door handle.

In \texttt{Interactive Gibson Environment}, the agent controls the mobile manipulator JackRabbot composed by a six degrees of freedom (DoF) Kinova arm mounted on a non-holonomic two-wheeled Segway mobile base. The agent commands actions for the mobile manipulator in continuous action space: desired velocities for each joint, including the two wheels and five joints of the arm (seven DoF in total). The observations to the agent include a depth map, and proprioceptive information such as the poses of the base and the end-effector, the joint configurations, and the goal position.

\vspace{-5pt}
\subsection{Experimental Setup}
\vspace{-5pt}
\label{ss:setup}

In our experiments, we use PPO~\cite{schulman2017proximal} as the policy gradient algorithm for both the high-level and the low-level policies of HRL4IN. When training both policies at the same time, the changing behavior of the low-level policy poses a non-stationary problem for the high-level policy: the same high-level action could result in different low-level actions, and therefore different state transitions and rewards, depending on the current training stage of the low-level policy. We believe an on-policy algorithm like PPO will stabilize the training process and encourage policy convergence. Details about the network structure and hyperparameters of HRL4IN and PPO can be found in Appendix~\ref{a:ns} and~\ref{appendix_training}.


To evaluate the performance of HRL4IN, we compare with two baselines: flat PPO and Hindsight Actor-Critic (HAC), a state-of-the-art HRL framework. Since the official implementation of HAC published by the authors does not support discrete action, we only compare HRL4IN with HAC in \texttt{Interactive Gibson Environment}. We run experiments in \texttt{Interactive Gibson Environment} and \texttt{Interactive ToyEnv} with 5 and 7 different random seeds, respectively. 

{\bf Interactive ToyEnv:}
The goal in \texttt{Interactive ToyEnv} is defined by the discrete coordinates of a desired agent's location, $g=(x,y)_{\mathit{WF}}\in \mathbb{N}^2$, where $\mathit{WF}$ indicates a global world frame. The agent starts at a random location in the left room with a random orientation. Subgoals are defined as the desired relative change in the agent's $x$, $y$ location, orientation and door state, $g_{t}^{\lowlevel} \in \mathbb{N}^4$. The embodiment selector can assume three values: \texttt{base-only}, \texttt{arm-only}, and \texttt{base+arm}. 

The reward is $r_{t} =  \mathbf{w}^T \mathbf{r}$, where the reward vector is defined as $\mathbf{r} = [r_{\mathit{success}}, r_{\mathit{energy}}]$ and the weights to sum them up are $\mathbf{w} = [w_{\mathit{success}}, w_{\mathit{energy}}]$. The reward weights are defined in Appendix~\ref{appendix_env}. In the reward vector, $r_{\mathit{success}} = \mathbbm{1}\{(x,y)_{t} = g\}$ indicates goal convergence, $r_{\mathit{energy}} = 0$ if there is no actuation, $r_{\mathit{energy}} = 1$ if there is arm \textbf{or} base actuation, and $r_{\mathit{energy}} = 2$ if there are base \textbf{and} arm actuations. An episode ends when the agent reaches the goal or runs out of time. We use time scale $T = 4$ for the high-level policy.

{\bf Interactive Gibson Environment:}
The goals in \texttt{Interactive Gibson Environment} is defined by the continuous coordinates of a desired agent's base location, $g=(x,y)_{\mathit{WF}}\in \mathbb{R}^2$. The agent starts at a fixed location with a random orientation and is assumed to converge to the goal when the distance between the agent base position and the goal is smaller than $t_{g} = \SI{0.5}{\meter}$. Subgoals $g_{t}^{\lowlevel}$ are defined as the desired relative change in the agent's end-effector position in either 3D or 2D (xy-plane) space, depending on the embodiment selection. The embodiment selector can assume two values: \texttt{base-only} and \texttt{base+arm}. The threshold to assume subgoal convergence is $t_{\mathit{sg}} = \SI{0.05}{\meter}$.

The environment returns different rewards in three different phases of the Interactive Navigation task: moving towards the door, opening the door, and moving towards to the goal. The reward is $r_{t} =  \mathbf{w}^T \mathbf{r}$, where the reward vector is defined as: $\mathbf{r}= [r_{\mathit{success}}, r_{\mathit{progress}}, r_{\mathit{energy}}, r_{\mathit{collision}}]$ and the weights to sum them up are $\mathbf{w} = [w_{\mathit{success}}, w_{\mathit{progress}}, w_{\mathit{energy}}, w_{\mathit{collision}}]$. The reward weights are defined in Appendix~\ref{appendix_env}. In the reward vector, $r_{\mathit{success}} = \mathbbm{1}\{|(x,y)_t - g|^2 < t_{g} \}$ indicates goal convergence; $r_{\mathit{progress}}$ is a small positive value if the agent is making progress for the current stage (e.g. getting closer to the door handle or opening the door partially); $r_{\mathit{energy}} = \mathbbm{1}\{\dot{q}_{\mathit{base}} > t_{\mathit{energy}}\} + \mathbbm{1}\{\dot{q}_{\mathit{arm}} > t_{\mathit{energy}}\}$, where $\dot{q}_{\mathit{base}}$ and $\dot{q}_{\mathit{arm}}$ are joint velocities of the base wheels and the arm joints; $r_{\mathit{collision}} = 1$ if the agent collides with the environment or itself. An episode ends when the agent reaches the goal, runs out of time or tips over. We use time scale $T = 50$ (equal to 5 seconds of world time) for the high-level policy.

\vspace{-5pt}
\subsection{Results}
\vspace{-5pt}
\label{ss:res}

\begin{figure}
\centering
\begin{subfigure}[b]{0.34\textwidth}
\includegraphics[width=0.99\textwidth]{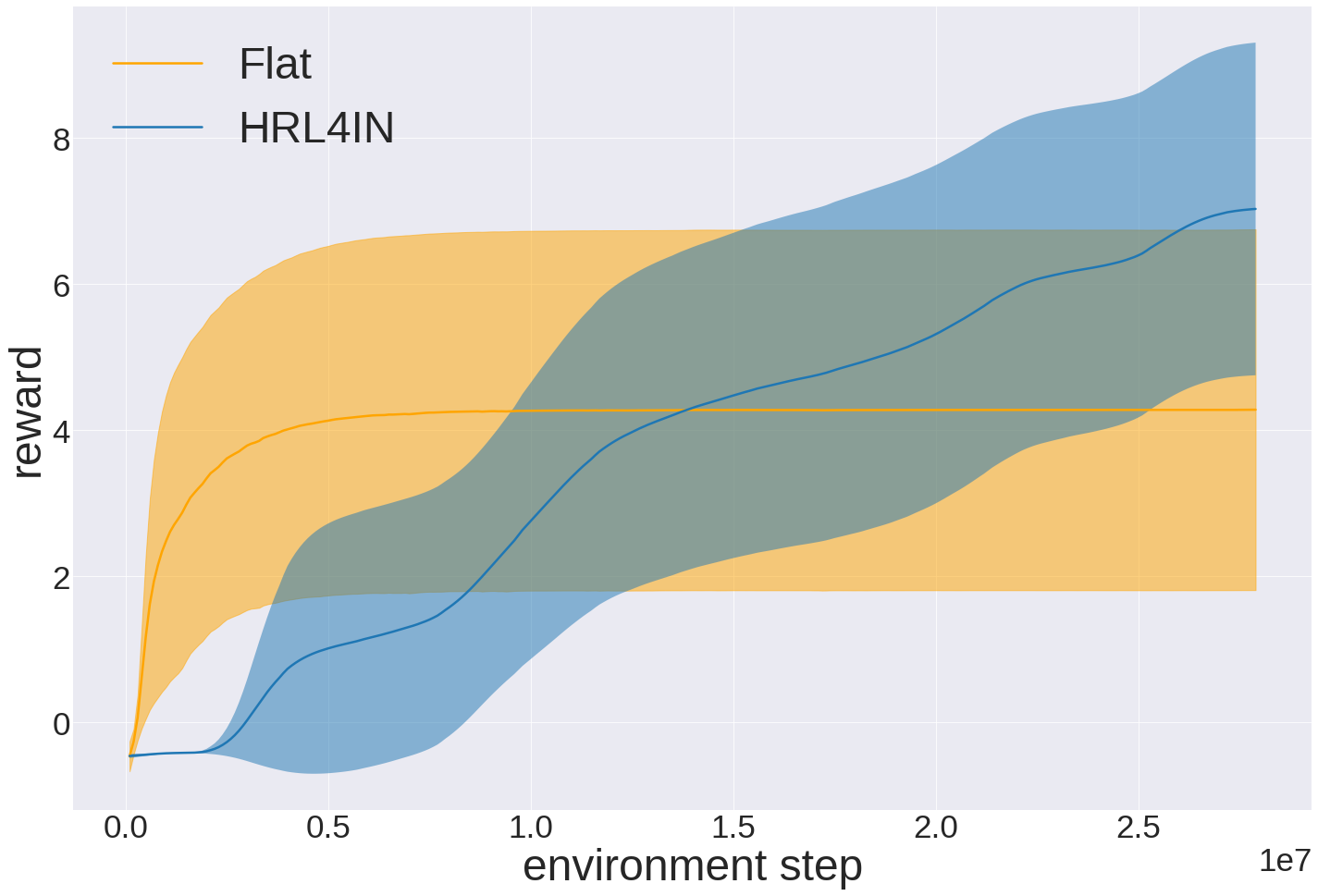}%
\caption{\small{Reward over time for HRL4IN and flat PPO; HRL4IN achieves higher reward across 7 random seeds despite being less sample efficient initially.}}
\label{sf:ite:rw}
\end{subfigure}%
\hfill
\begin{subfigure}[b]{0.64\textwidth}
\centerline{\includegraphics[width=0.85\textwidth]{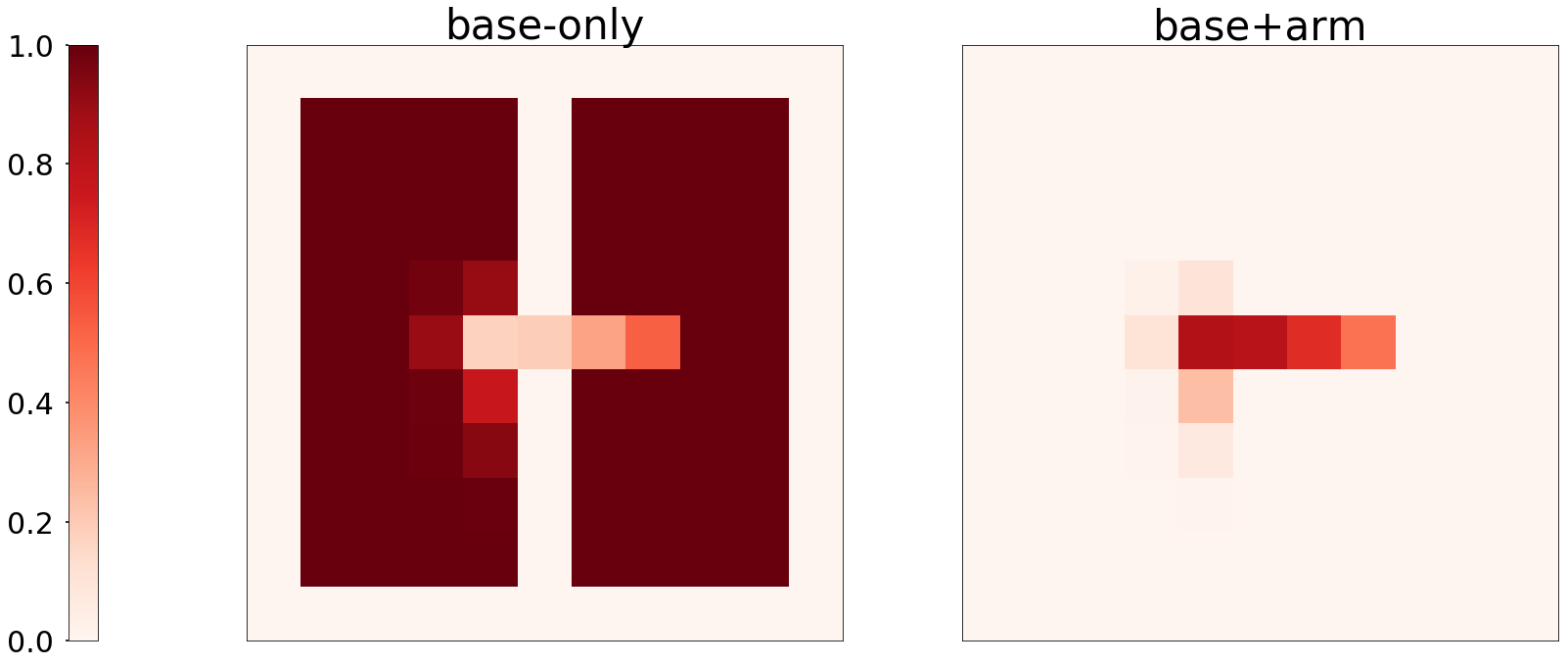}}%
\caption{\small{Probability of using different parts of the embodiment; the high-level policy learns to set base-only subgoals everywhere except when the agent is near the door, where it sets base+arm subgoals. Arm-only usage is excluded because it's overly restrictive and never used.}}
\label{sf:ite:emb}
\end{subfigure}%
\vspace{-10pt}
\caption{\small{Experimental results for \texttt{Interactive ToyEnv}}}
\label{f:ite:results}
\end{figure}

{\bf Interactive ToyEnv:} Our experimental results for the \texttt{Interactive ToyEnv} are depicted in Fig.~\ref{f:ite:results}. While being initially less sample efficient, HRL4IN achieves higher reward than flat PPO. HRL4IN learns to solve the task for 5 out of 7 random seeds while flat PPO only does so for 3 out of 7 random seeds. Given the sparse reward, HRL4IN is able to explore more efficiently than flat PPO through temporal extended commitments towards subgoals. The plot also shows that once flat PPO learns to solve the task, the policy consistently repeats the successful strategy. HRL4IN, on the other hand, has less stable training process since the same high-level action could result in different executions by the low-level policy that is being trained simultaneously. Our best performing model achieves 100\% success rate with average episode length of 19.2 steps. The optimal policy, computed analytically using the shortest path, takes on average 14 steps, so our solution is $<1.5\times$ optimal.

\begin{figure}
\centering
\includegraphics[width=0.25\textwidth]{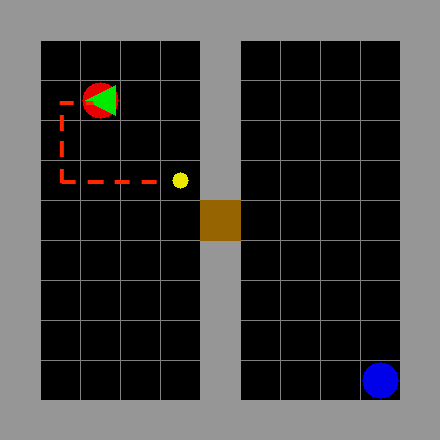}%
\vspace{10pt}
\includegraphics[width=0.25\textwidth]{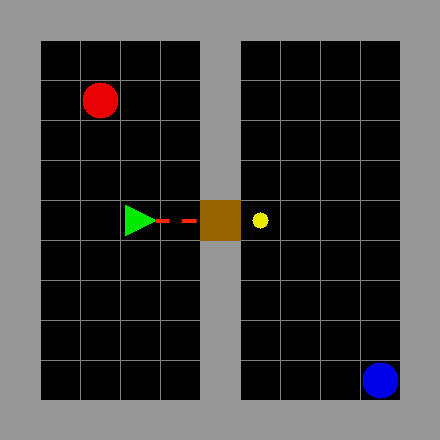}%
\vspace{10pt}
\includegraphics[width=0.25\textwidth]{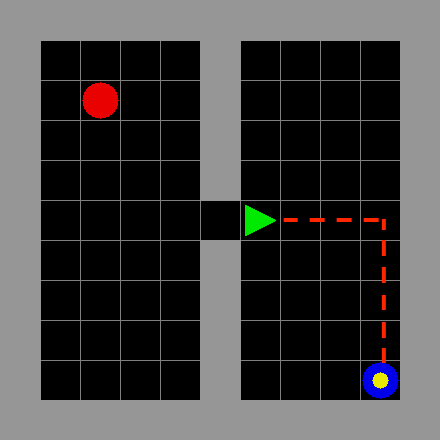}%
\caption{\small{Three phases of a sample trajectory for \texttt{Interactive ToyEnv}; the agent starts at a random location (red) in the left room and is rewarded for reaching the goal (blue) in the right room; (\textit{Left}) the agent first sets a base-only subgoal (yellow) to approach the door; (\textit{Middle}) the agent then sets a base+arm subgoal to get to the front of the door and to open it; (\textit{Right}) the agents sets a base-only subgoal to reach the final goal.}}
\vspace{-20pt}
\label{f:ite:execution}
\end{figure}

We also analyze the probability of using different parts of the embodiment at different locations of the rooms (Fig.~\ref{sf:ite:emb}). We observe that the high-level policy of HRL4IN learns to use only base at almost all locations except those near the door, where it uses both base and arm. It is energy-efficient to use only base when far away from the door because the agent cannot interact with it. The agent never uses arm-only because unless the agent is 1) at the position in front of the door and 2) facing it, it needs to use both base and arm to make progress towards the goal and arm-only is overall too restrictive. Fig.~\ref{f:ite:execution} illustrates three different phases of a sample trajectory of our best model.


\begin{figure}
\centering
\begin{subfigure}[b]{0.45\textwidth}
\includegraphics[width=\textwidth]{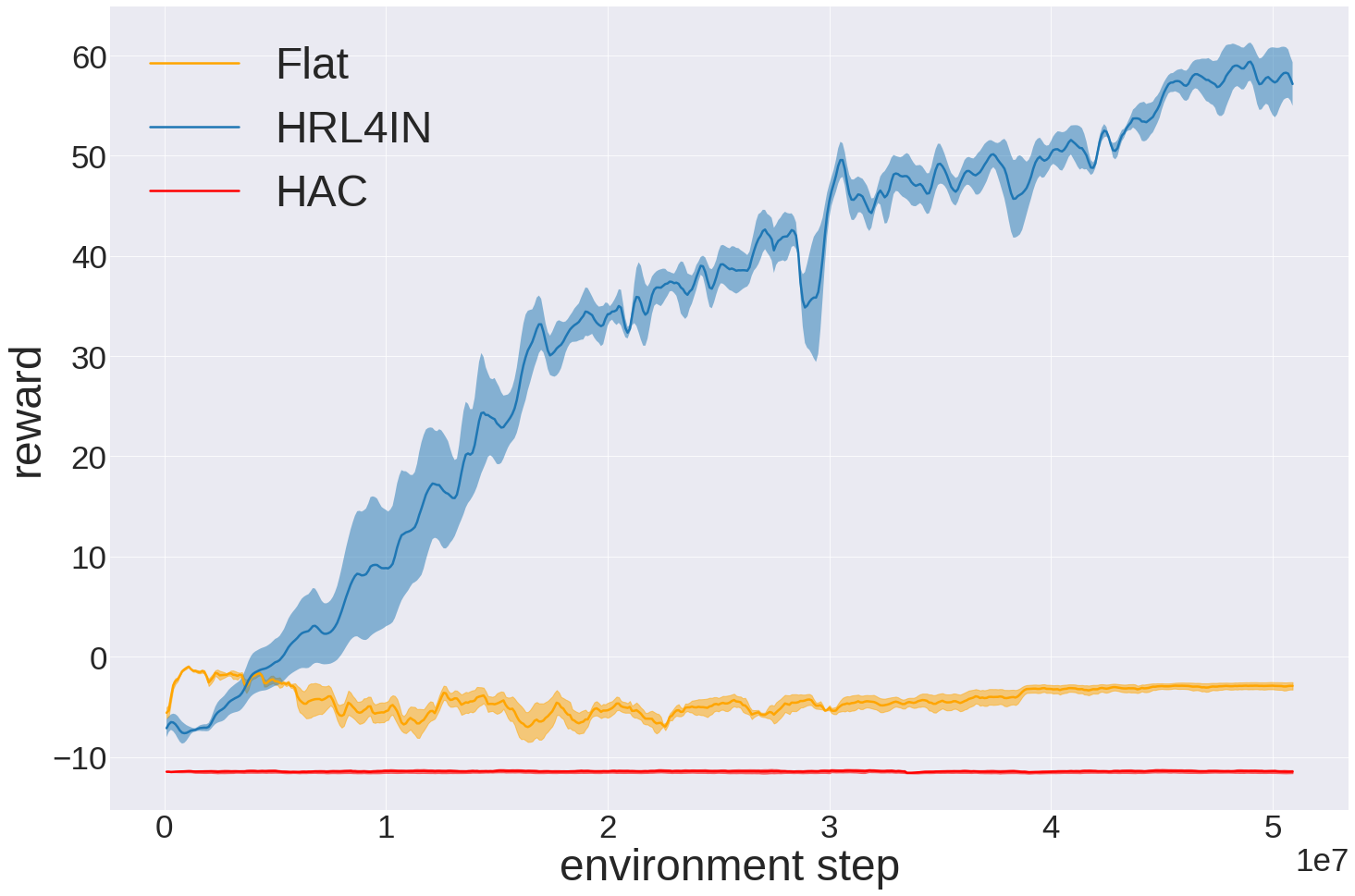}%
\caption{\small{Reward over time for HRL4IN, flat PPO and HAC; HRL4IN significantly outperforms both baselines in the final reward, indicating strong exploration capabilities of our algorithm.}}
\label{sf:ige:rw}
\end{subfigure}%
~~~~
\begin{subfigure}[b]{0.58\textwidth}
\centering
\hspace*{3pt}
\includegraphics[width=0.275\textwidth]{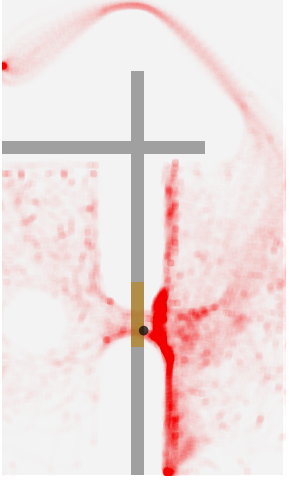}
\includegraphics[width=0.275\textwidth]{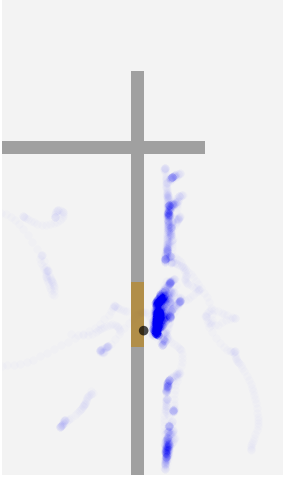}%
\\
\begin{minipage}{0.3\linewidth}
\begin{center} \scriptsize{~~~~~~~Only Base Usage} \end{center}
\end{minipage}
\begin{minipage}{0.3\linewidth}
\begin{center} \scriptsize{Base and Arm Usage} \end{center} 
\end{minipage}
\caption{\small{Scatter plot of embodiment selections at different locations of 100 trajectories; the high-level policy learns to set base-only subgoals at most of the locations to save energy and set base-arm subgoals when in front of the door to grasp the door handle.}}
\label{sf:ige:emb}
\end{subfigure}%
\caption{\small{Experimental results for \texttt{Interactive Gibson Environment}. }}
\vspace{-5pt}
\label{f:ige}
\end{figure}

{\bf Interactive Gibson Environment:} Our experimental results for the \texttt{Interactive Gibson Environment} are depicted in Fig.~\ref{f:ige}. Similar to the pattern observed for the \texttt{Interactive ToyEnv}, the reward increases faster initially for flat PPO than for HRL4IN: flat PPO quickly starts to exploit the experiences from the initial exploration and gets stuck in a local maximum of return that corresponds to the scenario in which the agent moves towards the wall separating the initial room and the final room (closer to the final goal). On the contrary, HRL4IN performs a deeper exploration in subgoal directions and achieves much higher reward in the long run. The other baseline HAC completely fails to solve the task. Our hypothesis for HAC’s poor performance is that our task poses a difficult exploration problem in which there exists a discrepancy between the final goal distribution and the hindsight goal distribution. The subgoals that HAC learns to achieve during training does not contribute to achieving the final goal. We observe that most transitions sampled from the replay buffer include hindsight goals very close to the initial position and high virtual rewards, and few include real goals and low real rewards. Our best performing HRL4IN policies has close to 100\% success rate after 50M training steps. A few failure cases are caused by the agent accidentally colliding with the door and closing it back on its way to the final goal.

We have run ablation studies to understand the effect of embodiment selection and different reward terms in \texttt{Interactive Gibson Environment} on model performance (Table \ref{t:ablation_studies}). First, we ablate the embodiment selector for HRL4IN and find that it achieves comparable reward and success rate, but fails to save any energy. We compute energy in a symbolic sense since we can't accurately measure energy consumption from the simulator: the agent consumes 2 units of energy when using base and arm, and 1 unit of energy when using only base or arm. Furthermore, we ablate each reward term except $r_{\mathit{success}}$ and observe that all reward terms are necessary for training a successful, energy-efficient policy: without $r_{\mathit{progress}}$ and $r_{\mathit{collision}}$, the agent fails to solve the task; without $r_{\mathit{energy}}$, the agent fails to save energy or learn meaningful embodiment selection.

\begin{figure}[!t]
\begin{subfigure}[t]{0.24\textwidth}
\centering
\includegraphics[height=2.5cm]{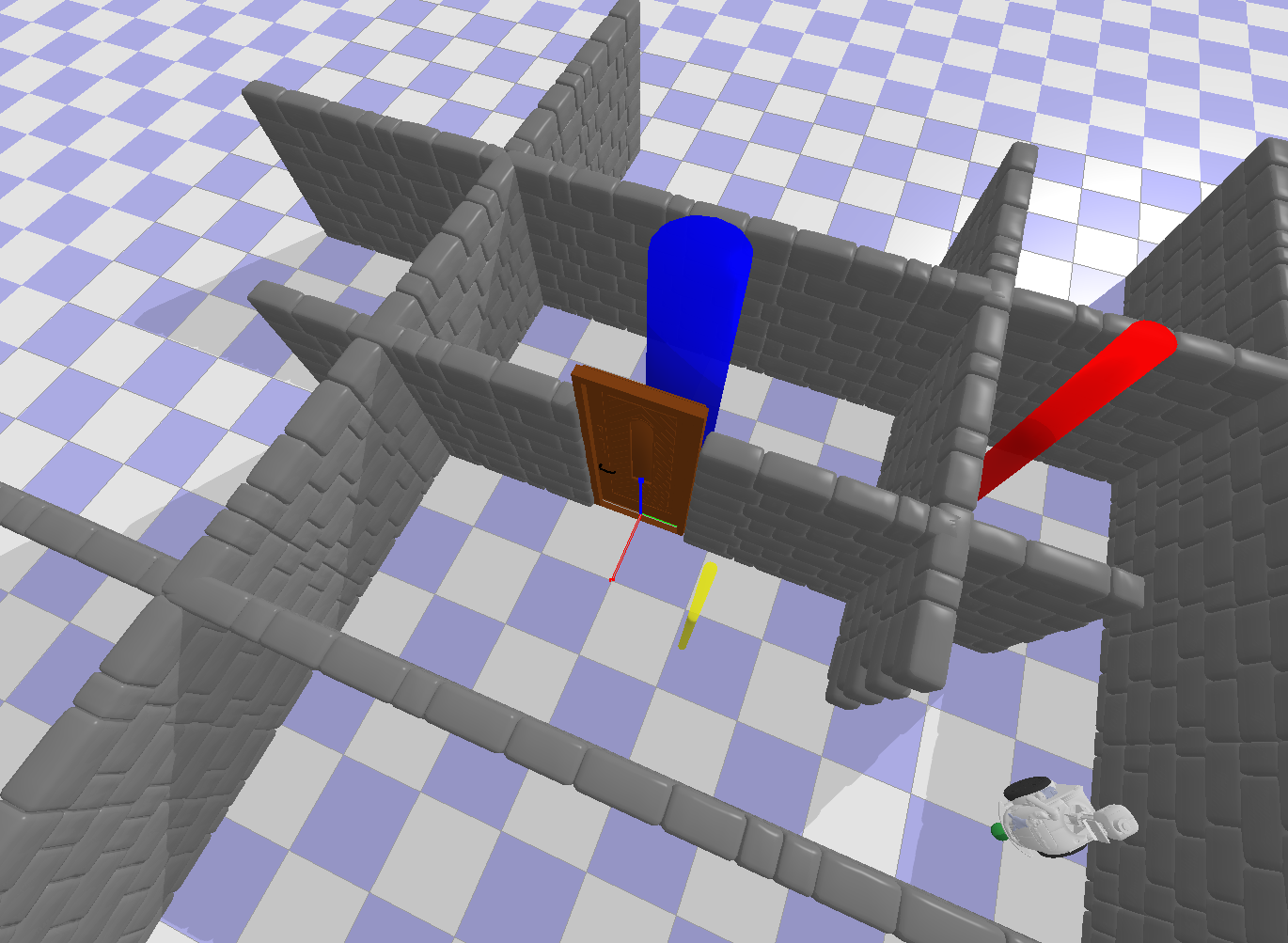}%
\caption{}
\end{subfigure}
~\begin{subfigure}[t]{0.23\textwidth}
\centering
\includegraphics[height=2.5cm]{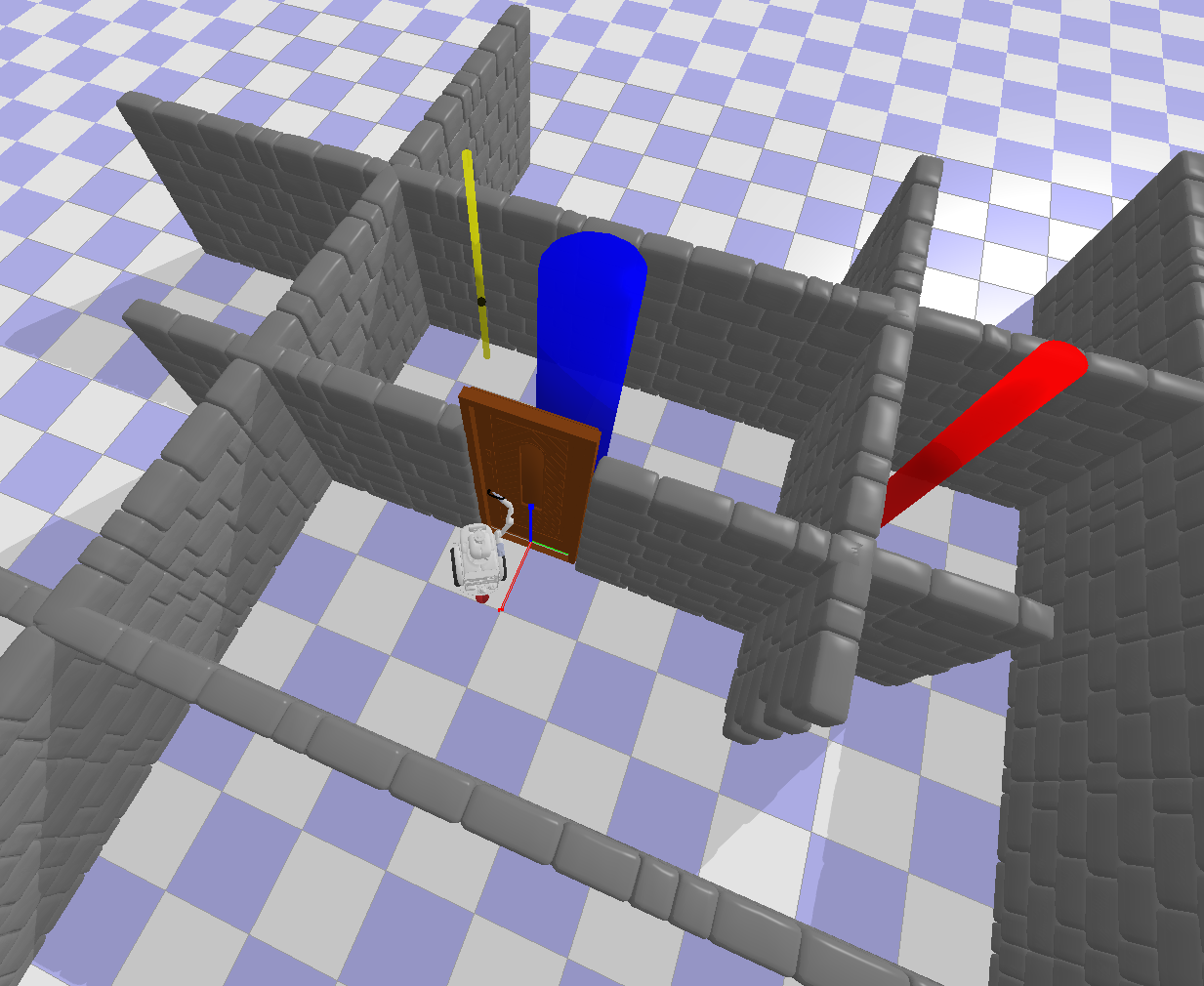}%
\caption{}
\end{subfigure}
~\begin{subfigure}[t]{0.25\textwidth}
\centering
\includegraphics[height=2.5cm]{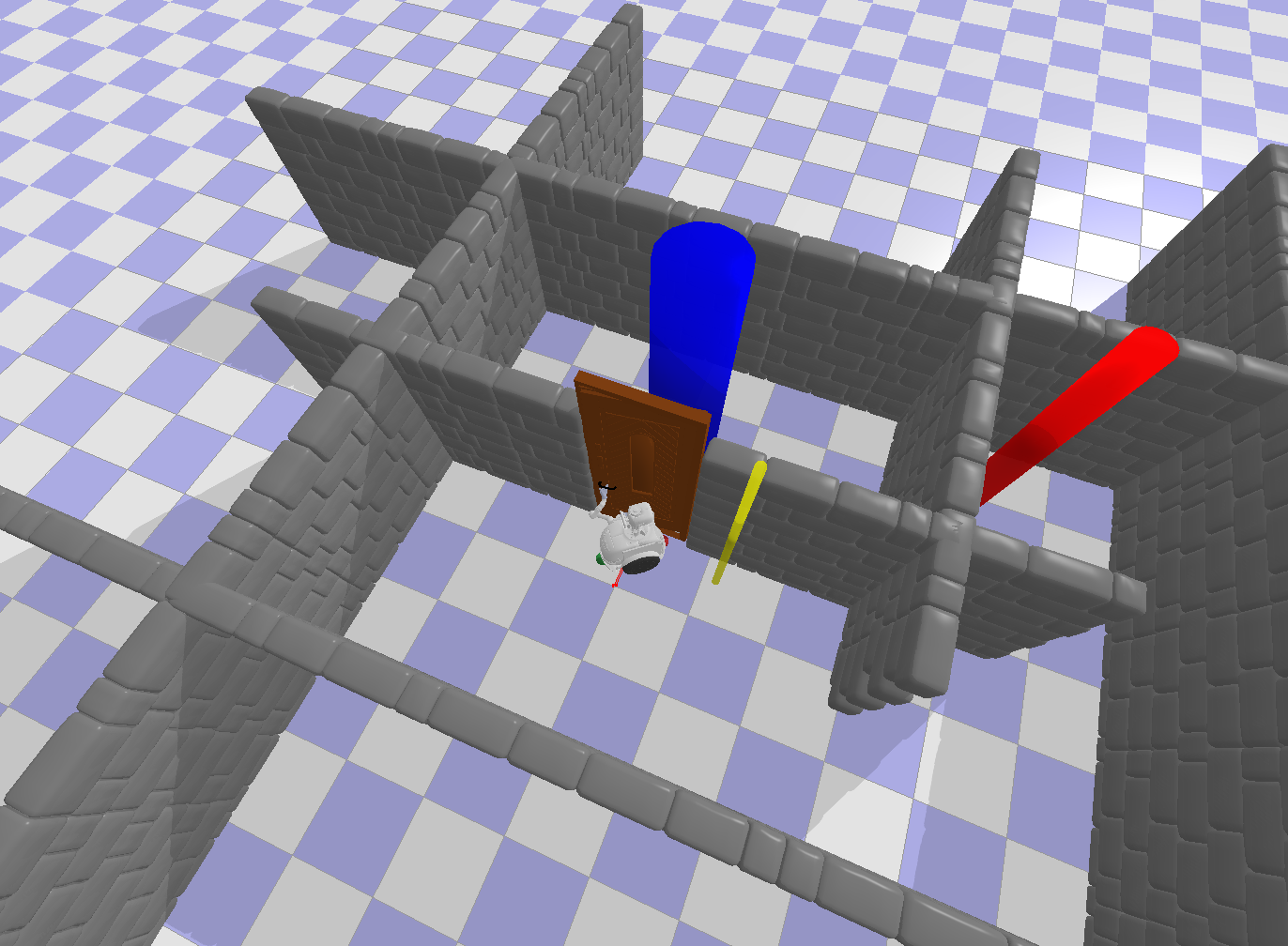}%
\caption{}
\end{subfigure}
~\begin{subfigure}[t]{0.24\textwidth}
\centering
\includegraphics[height=2.5cm]{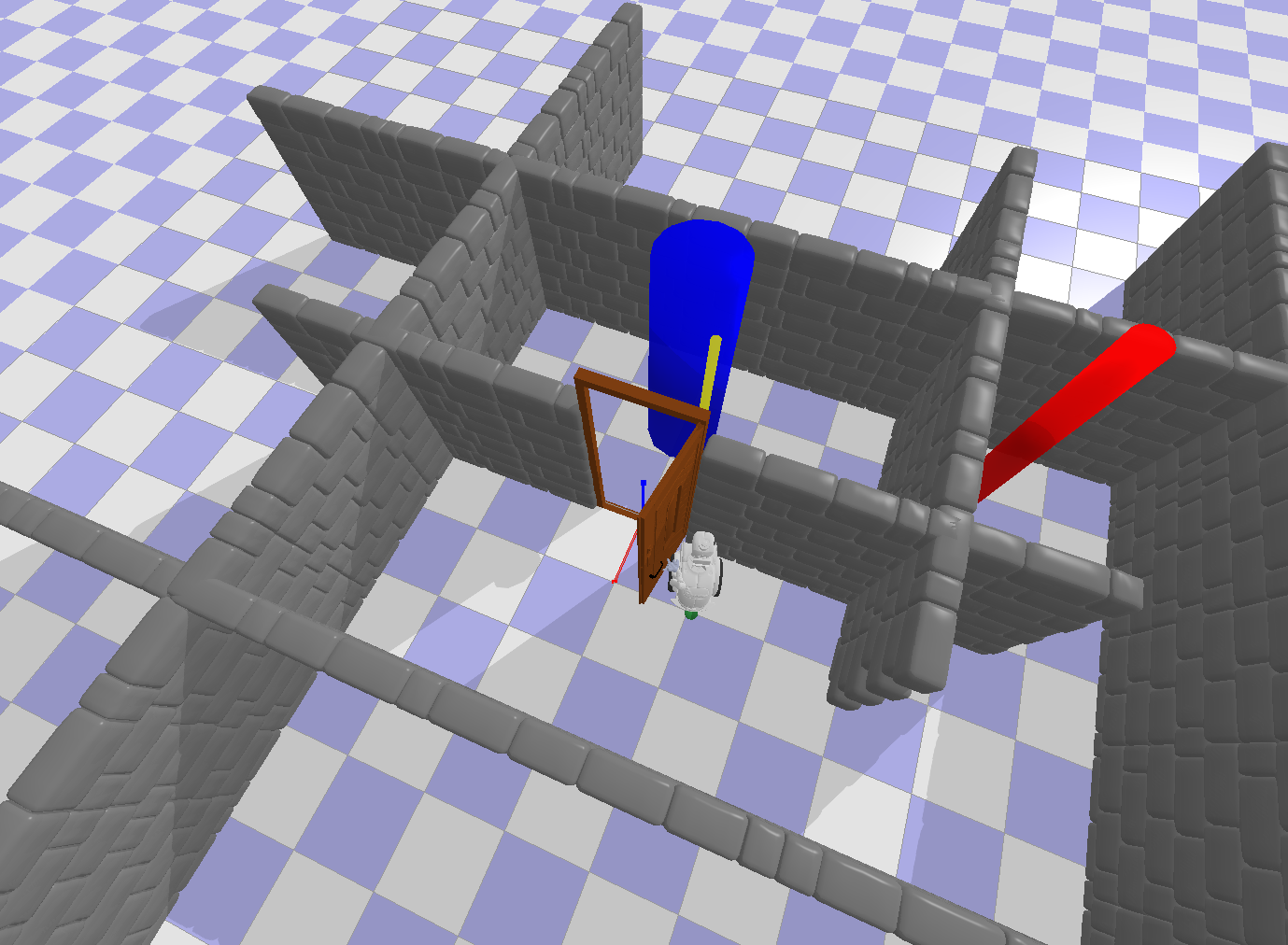}%
\caption{}
\end{subfigure}
\caption{\small{Four phases of a sample trajectory for \texttt{Interactive Gibson Environment}; (a) the agent navigates from the initial location (red cylinder) towards the door by following a base-only subgoal (yellow cylinder); (b) when in front of the door, the agent grabs the door handle with the arm following a base+arm subgoal (black sphere in the yellow cylinder); (c) the agent moves its base to open the door by setting a base-only subgoal behind itself (d) the agent follows a base-only subgoal to reach the final goal (blue cylinder).}}
\label{f:ige:phases}
\vspace{-10pt}
\end{figure}

We also analyze the use of different parts of the embodiment at different locations of the environment (Fig.~\ref{sf:ige:emb}). We plot the embodiment selections from 100 evaluation episodes on the room layout. We observe that the high-level policy learns to use only base most of the time to save energy. It sometimes uses base and arm when in front of the door to grasp the door handle. Then it tends to switch back to use only base again because now that the end-effector is attached to the door handle, moving the base is sufficient to open the door. Finally, it uses only base in the final room since the door is already open. Fig.~\ref{f:ige:phases} illustrates four different phases of a sample trajectory of our best model. More experimental results and policy visualizations can be found in Appendix~\ref{appendix_additional_results}.



\begin{table}
\centering
\small
\caption{\small{Ablation Study Results for \texttt{Interactive Gibson Environment}. We show here all reward terms and embodiment selection are necessary to achieve the best success rate and energy-saving.}}
\begin{tabular}{ccccc}
\toprule
Model & Reward Terms & Success Rate & Reward & Energy-Saving \\ \midrule
\bf{HRL4IN (full)} & \bf{All} & {$\mathbf{0.963}$} &  {$\mathbf{64.3}$} & $\mathbf{0.453}$ \\ 
HRL4IN (w/o emb. sel.) & All & $0.912$ & $61.5$ & $0.0$ \\ 
HRL4IN (full) & No $r_{\mathit{energy}}$ & $0.934$ & $63.4$ & $0.235$ \\ 
HRL4IN (full) & No $r_{\mathit{collision}}$ & $0.0$ & $9.0$ & $0.421$ \\ 
HRL4IN (full) & No $r_{\mathit{progress}}$ & $0.0$ & $-1.2$ & $0.498$ \\ \bottomrule
\end{tabular}
\label{t:ablation_studies}
\end{table}

\normalsize

\vspace{-5pt}
\section{Conclusion}
\label{s:con}
\vspace{-5pt}

We presented HRL4IN, a hierarchical Reinforcement Learning solution for Interactive Navigation tasks, tasks that require motion of the base and the arm of a mobile manipulator. The high-level policy of HRL4IN sets subgoals in different subspaces (navigation and/or manipulation) for the low-level policy to achieve. Moreover, the high level can also decide on the part of the embodiment (base, arm, base and arm) to use at each temporally extended phase to perform deep exploration. We evaluate HRL4IN against flat PPO and HAC in two environments, a grid-world environment with discrete action space, and a complex, realistic robotics environment with continuous action space. Our results indicate that HRL4IN achieves higher reward and solve the tasks more successfully and efficiently. In future work, we would like to 1) examine other relevant tasks that require interactive navigation and 2) transfer the learned policy to the real world. 

\acknowledgments{We acknowledge the support of Toyota Research Institute (``TRI'') but this article solely reflects the opinions and conclusions of its authors and not TRI or any other Toyota entity. Fei Xia would like to acknowledge the support of Stanford Graduate Fellowship.}

\clearpage
\footnotesize
\bibliography{main}  

\clearpage

\normalsize
\begin{appendices}
\section{Environment Details}
\label{appendix_env}

{\bf Interactive ToyEnv:}
We use $k = 11$ for our grid size. The room configuration is depicted in Fig.~\ref{fig:environments}. The door state has a minimum value of 1 (closed) and a maximum value of 5 (open). The agent will increment the door state by 1 if it outputs \texttt{slide-up} at the front of the door while facing it and decrease by 1 if it outputs \texttt{slide-down}. The maximum episode length is 500.

The full observation includes:
\begin{itemize}[nosep]
  \item agent position
  \item agent orientation represented by yaw
  \item door state
  \item cosine and sine of yaw
  \item goal position
  \item whether the agent is next to the door
  \item a top-down global map with four channels: the first channel encodes the static environment, the second channel encodes agent's position and orientation, the third channel encodes the goal position and the last channel encodes the door position and state.
\end{itemize}

The high-level policy only sets subgoals for the first three elements of the observation because the other elements are either immutable or redundant information that solely helps the neural networks to learn policies more efficiently.

The reward weights defined in Sec.~\ref{ss:setup} are $w_{success}= 10.0$ and $w_{energy}=-0.001$.


{\bf Interactive Gibson Environment:}
We used a room size of approximately $\SI{6}{\meter} \times \SI{9}{\meter}$. The room configuration is depicted in Fig.~\ref{fig:environments}. The door is considered fully open if the door angle is more than \ang{90}. The maximum episode length is $1000$.

\begin{table}[b]
\begin{subfigure}[t]{0.45\textwidth}
\centering
\scriptsize
\begin{tabular}{lc}
\toprule
Weight & Value \\ \midrule
$w_{\mathit{success}}$ & $30.0$ \\ 
$w_{\mathit{progress}}$ & $2.0$ \\  
$w_{\mathit{energy}}$ & $-0.001$ \\ 
$w_{\mathit{collision}}$ & $-0.01$ \\ \bottomrule
\end{tabular}
\caption{Reward weights}
\label{t:igw}
\end{subfigure}
\hfill
\begin{subfigure}[t]{0.45\textwidth}
\centering
\begin{center}
\scriptsize
\begin{tabular}{lc}
\toprule
Parameter & Value \\ \midrule
Simulation time step & $\SI{1/40}{\second}$ \\ 
Action timestep: & $\SI{1/10}{\second}$ \\ 
Lateral friction coefficient & $1.0$\\ 
Rolling friction coefficient &  $0.0$\\ \bottomrule
\end{tabular}
\end{center}
\caption{Physics simulation parameters}
\label{t:physics}
\end{subfigure}
\caption{Parameters for \texttt{Interactive Gibson Environment}}
\end{table}

The full observation includes:
\begin{itemize}[nosep]
  \item end-effector position $(x, y, z)_{\mathit{ee}}$
  \item base position
  \item linear and angular velocity
  \item roll, pitch, yaw of the base
  \item joint space configuration, velocities and torques for wheels and arm joints
  \item door angle and location
  \item goal location
  \item a flag that indicates collision or self-collision
  \item a flat that indicates if the end-effector is currently grabbing to the door handle
  \item cosine and sine of the robot's base orientation, joint configurations for revolute joints, and door angle
  \item a $64\times64$ depth map clipped to a range of $\SI{5}{\meter}$
\end{itemize}

For similar reasons as in the \texttt{Interactive ToyEnv}, the high-level policy only sets subgoals for the first element of the observation vector, the end-effector position $(x, y, z)_{\mathit{ee}}$. When the high level selects to use only the base embodiment ($e_{t'}=$\texttt{base-only}), the $z$ coordinate is \textit{masked out} for the computation of intrinsic reward by the associated subgoal mask and the desired velocities for all joints of the arm are set to zero by the associated action mask. The reward weights defined in Sec.\ref{ss:setup} are included in Table~\ref{t:igw}. The physics simulation parameters are included in Table~\ref{t:physics}.

\section{Method Details} 
\subsection{HRL4IN Algorithm}
\label{appendix_algo}

A detailed pseudo-code of our HRL4IN solution is included below in Algorithm~\ref{hrl4in_algo}:

\begin{algorithm}[H]
\label{hrl4in_algo}
\DontPrintSemicolon
\SetAlgoLined
\SetKwInOut{Input}{Environment}\SetKwInOut{Output}{Output}
\SetKwInOut{Parameters}{Parameters}
\Input{\texttt{env}}
\Output{$\pi_{\highlevel}$, $\pi_{\lowlevel}$}
\Parameters{$n_{update}, n_{step}, t_{sg}$}
\BlankLine
 
$n_{subgoal} \gets 0$ \\
$R_{extrinsic} \gets 0$ \\
\For{$update \gets 0$ \KwTo $n_{update}$}{
   \For{$step \gets 0$ \KwTo $n_{step}$}{
    
     \uIf{$n_{subgoal} = 0$}{
        $g^{\lowlevel}, e \gets \pi_{\highlevel}(obs)$ \tcp*{high level gives subgoal and embodiment type} 
        $m^{act}, m^{sg} \gets get\_masks(e)$ \\
        \tcp*{determine action and subgoal mask from embodiment type}
        $g_{cached}^{\lowlevel} \gets g^{\lowlevel}$ \\
        $obs_{cached} \gets obs $\\
        $g_{abs}^{\lowlevel} \gets g^{\lowlevel} + obs$ \tcp*{convert subgoal from relative to absolute}
    }
    
    $a_{\lowlevel} \gets \pi_{\lowlevel}(obs, g^{\lowlevel}, m^{act}, m^{sg})$ \\
    $a \gets a_{\lowlevel} \otimes m^{act}$ \tcp*{apply action mask}
    $next\_obs ,r \gets \texttt{env.step}(a)$ \tcp*{collect experience from environment}
    $n_{subgoal} \gets n_{subgoal} + 1$ \\
    $R_{extrinsic} \gets R_{extrinsic} + r $  \tcp*{accumulate extrinsic reward}
    
      \uIf{$||next\_obs - g_{abs}^{\lowlevel}||_2 < t_{sg}$ \text{or} $n_{subgoal} = T$ \tcp*{subgoal achieved or time out}} { 
      $ReplayBuffer_{\highlevel} \gets (obs_{cached}, g_{cached}^{\lowlevel}, R_{extrinsic}, next\_obs)$ \\
      $n_{subgoal} \gets 0$ \\
      $R_{extrinsic} \gets 0$
      }

    $R_{intrinsic} \gets ||(obs - g_{abs}^{\lowlevel})\otimes m^{sg}||_2 - ||(next\_obs - g_{abs}^{\lowlevel})\otimes m^{sg}||_2$ \\ 
    \tcp*{calculate intrinsic reward}
    $ReplayBuffer_{\lowlevel} \gets (obs, a_{\lowlevel}, R_{intrinsic}, next\_obs)$ \\
    $g^{\lowlevel} \gets g_{abs}^{\lowlevel} - next\_obs$ \tcp*{convert subgoal from absolute to relative} 
    $n_{subgoal} \gets n_{subgoal} + 1$ \\
    $obs \gets next\_obs$ 
    }
    $ReplayBuffer_{\highlevel}\texttt{.compute\_advantage\_estimates()}$ \\
    \tcp*{subroutine as defined in \cite{schulman2017proximal}}
    $ReplayBuffer_{\lowlevel}\texttt{.compute\_advantage\_estimates()}$ \\
    $\pi_{\highlevel}\texttt{.update}(ReplayBuffer_{\highlevel})$ \tcp*{subroutine as defined in \cite{schulman2017proximal}}
    $\pi_{\lowlevel}\texttt{.update}(ReplayBuffer_{\lowlevel})$ \\
    $ReplayBuffer_{\highlevel}\texttt{.clear()}$ \\
    $ReplayBuffer_{\lowlevel}\texttt{.clear()}$ 
}

\caption{HRL4IN Algorithm}
\label{alg}
\end{algorithm}

\subsection{Network structure}
\label{a:ns}
We implement the high-level and low-level policies as Gated Recurrent Neural Networks (GRU~\cite{chung2014empirical}). For feature extraction, the observation space that contains spatial information (e.g. global map in \texttt{Interactive ToyEnv} and depth map in \texttt{Interactive Gibson Environment}) are processed by three convolutional layers and one fully connected layer with ReLU activation after flattening. Other observations are concatenated together and processed by one fully connected layer with ReLU activation. Finally, the features from two branches are concatenated together and fed into the GRU. The output features from the GRU are used to estimate the value function and the action distribution. Note that the embodiment selection component of the high-level action is modeled as a discrete action (e.g. choose between base-only and base-arm). The action losses of the subgoal and the embodiment selector are aggregated together for back-propagation. We follow the same training procedure as PPO~\cite{schulman2017proximal} and the training hyper-parameters can be found below.

\subsection{Training Details and Hyperparameters}
\label{appendix_training}
Table~\ref{t:hhrl} summarizes the hyperparameters for HRL4IN.

\begin{table}[!h]
\begin{center}
\scriptsize
\caption{Hyperparameters for HRL4IN}
\label{t:hhrl}
\begin{tabular}{lc}
\toprule
Hyperparameter & Value \\ \midrule
Learning Rate for High-Level Policy & $0.00001$ \\
Learning Rate for Low-Level Policy & $0.0001$ \\ 
Freeze High-Level Policy for First N Updates & $500$ \\ 
Intrinsic Reward Scaling & $30.0$ \\ 
Discount Factor for the High-Level policy & $0.99$ \\ 
Discount Factor for the Low-Level policy & $0.99$ \\ 
Subgoal Init. Std. Dev. (Gibson Env) & $[\SI{0.8}{\meter}, \SI{0.8}{\meter}, \SI{0.1}{\meter}]$ \\ 
Subgoal Min. Std. Dev. (Gibson Env) & $[\SI{0.05}{\meter}, \SI{0.05}{\meter}, \SI{0.05}{\meter}]$ \\ 
Subgoal Init. Std. Dev. (Toy Env) & $[0.2, 0.2, 0.5, 0.5]$ \\
Subgoal Min. Std. Dev. (Toy Env) & $[0.1, 0.1, 0.25, 0.25]$ \\ \bottomrule
\end{tabular}
\end{center}
\end{table}

Table~\ref{t:hppo} summarizes the hyperparameters for PPO.

\begin{table}[!h]
\begin{center}
\scriptsize
\caption{Hyperparameters for PPO}
\label{t:hppo}
\begin{tabular}{ l c }
\toprule
Hyperparameter & Value \\ \midrule
Learning Rate & $0.0001$ \\ 
PPO Clip Parameter & $0.2$ \\ 
PPO Epochs & $4$ \\
Num of Mini-Batches for PPO & $8$ \\
Value Loss Coefficient & $0.5$ \\ 
Entropy Coefficient & $0.01$ \\ 
Max Gradient Norm & $0.5$ \\ 
Rollout Steps & $256$ \\ 
Hidden layer size for the GRUs & $512$ \\ 
Use Linear LR Decay & True \\ 
Use Generalized Advantage Estimation (GAE) & True \\ 
Tau for GAE & 0.95 \\
Primitive Action Init. Std. Dev. & 0.33 \\ 
Primitive Action Min. Std. Dev. & 0.1 \\ \bottomrule
\end{tabular}
\end{center}
\end{table}

We normalize the observation space and the action space to $[-1, 1]$ during training and inference.

\section{Additional Results}
\label{appendix_additional_results}
In this section we show some additional results from the experiments described in Sec~\ref{s:exp}.

\begin{figure}
\centering
\begin{subfigure}[b]{0.50\textwidth}
\includegraphics[width=0.99\textwidth]{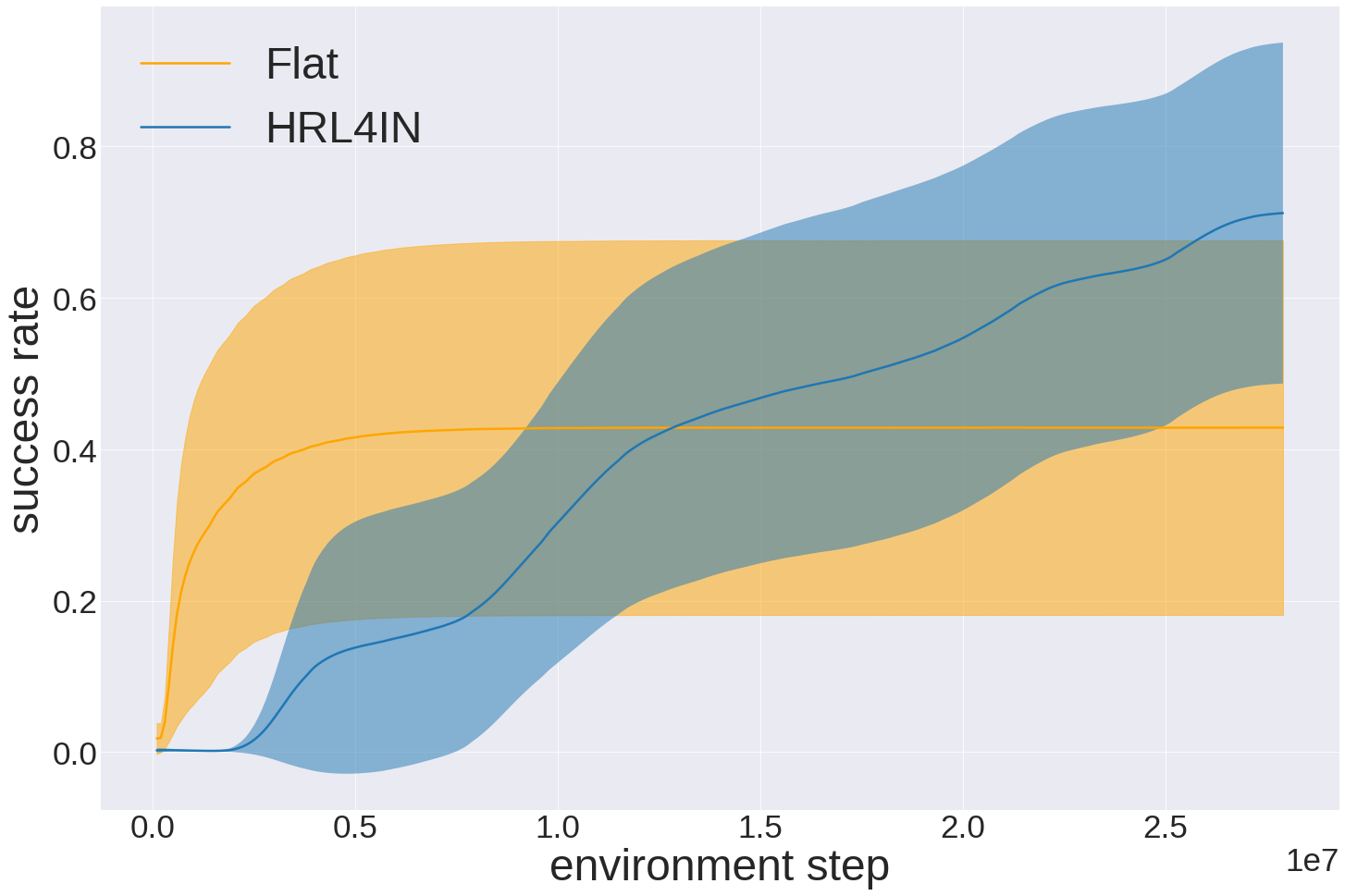}%
\label{sf:toy_succ}
\caption{}
\end{subfigure}%
~\begin{subfigure}[b]{0.50\textwidth}
\includegraphics[width=0.99\textwidth]{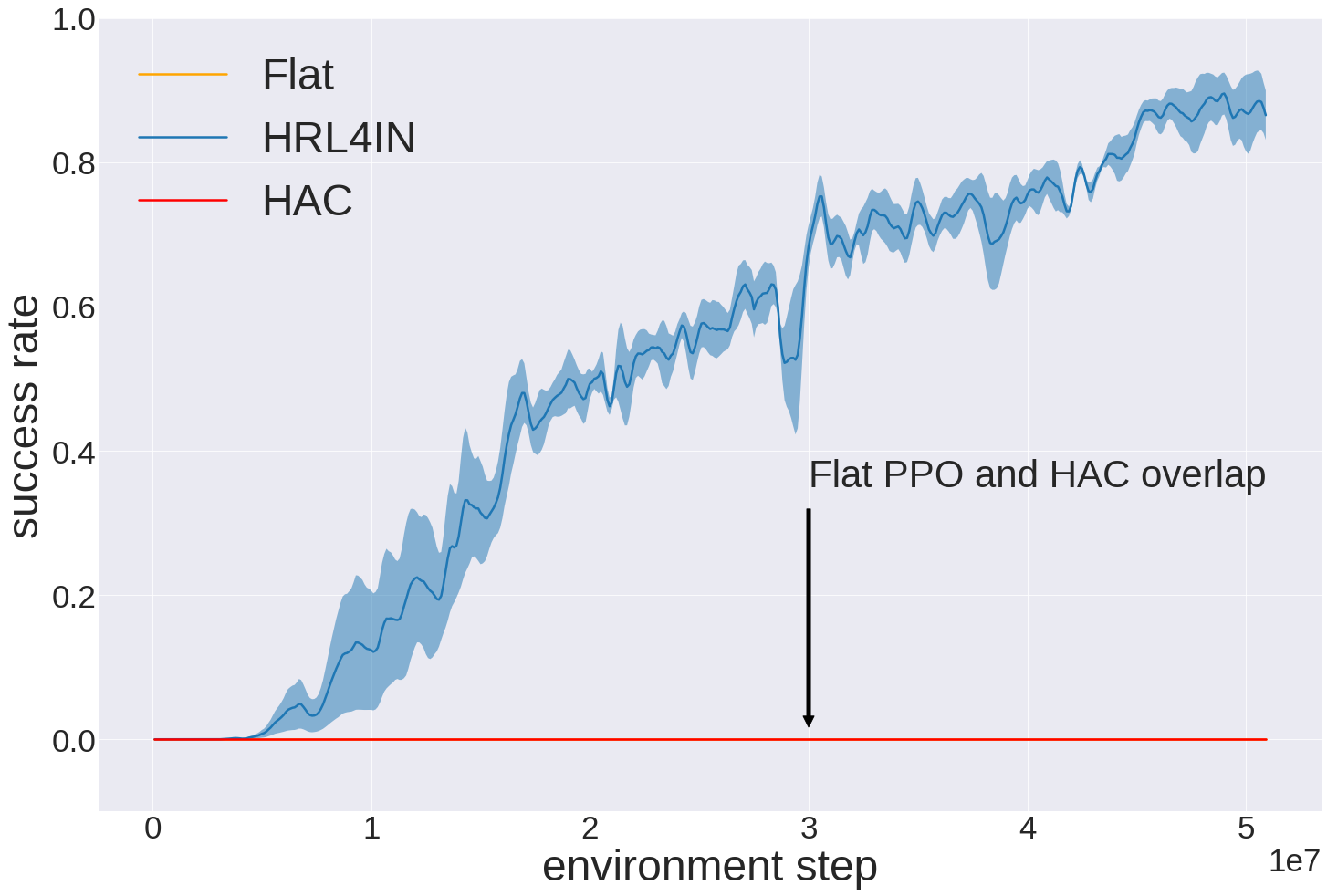}%
\label{sf:gibson_succ}
\caption{}
\end{subfigure}%
\caption{(a) Success rate over time for HRL4IN and flat PPO in \texttt{Interactive ToyEnv}, HRL4IN achieves higher success rate despite requiring more samples during the initial phase of training; (b) Success rate over time for HRL4IN, flat PPO and HAC in \texttt{Interactive Gibson Environment}; HRL4IN achieves close to 100\% success rate while the two baselines fail to solve the task.}
\label{fig:add_res}
\end{figure}

Fig.~\ref{fig:add_res} shows additional training curves of success rate on \texttt{Interactive ToyEnv} and \texttt{Interactive Gibson Environment}, which complements Fig.~\ref{sf:ite:rw} and Fig.~\ref{sf:ige:rw}. It shows that our approach is able to achieve higher final success rate against both baselines in both environments.

\begin{figure}
\centering
\begin{subfigure}[b]{0.33\textwidth}
\includegraphics[width=0.99\textwidth]{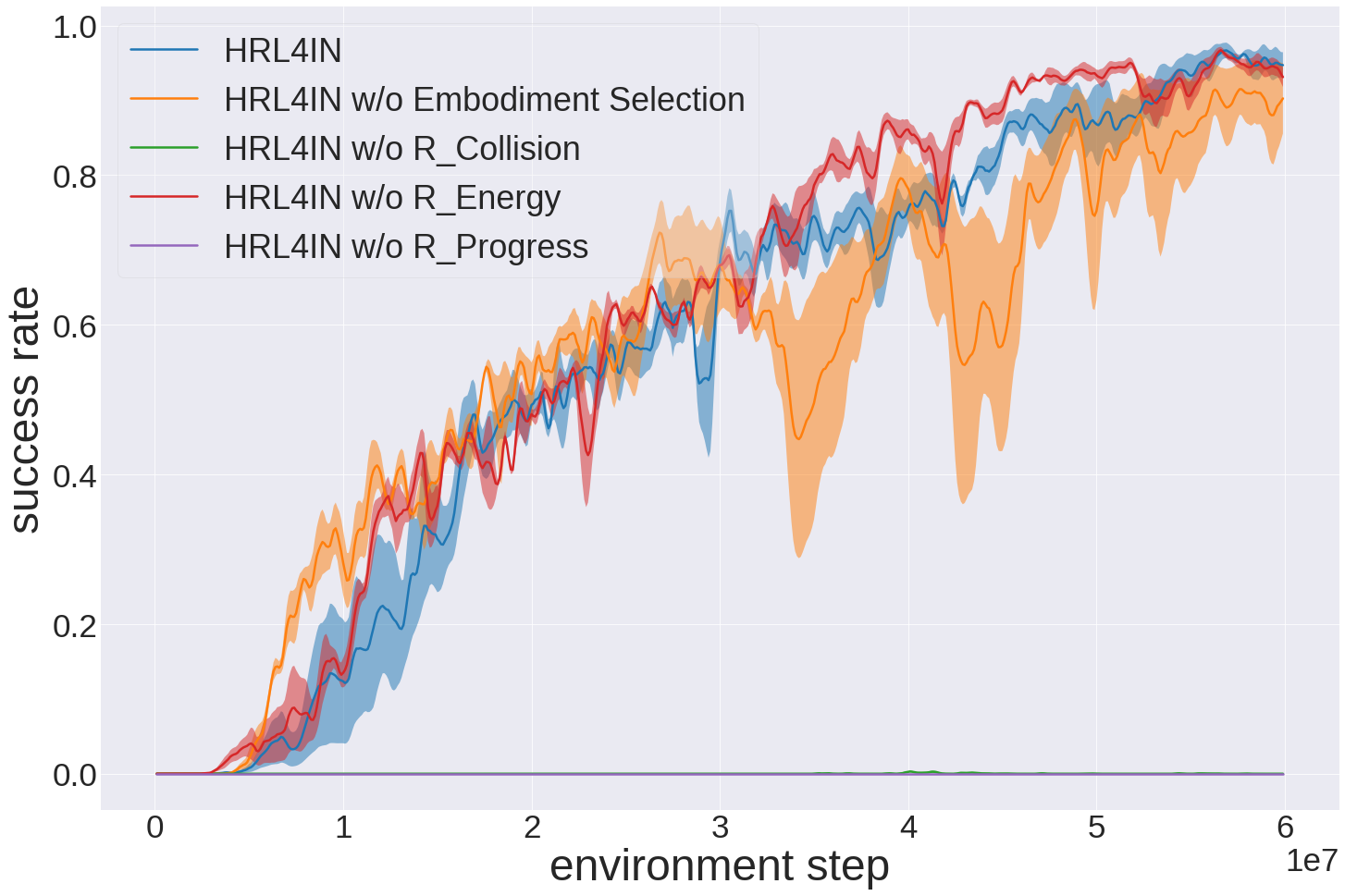}%
\label{sf:toy_succ}
\caption{}
\end{subfigure}%
~\begin{subfigure}[b]{0.33\textwidth}
\includegraphics[width=0.99\textwidth]{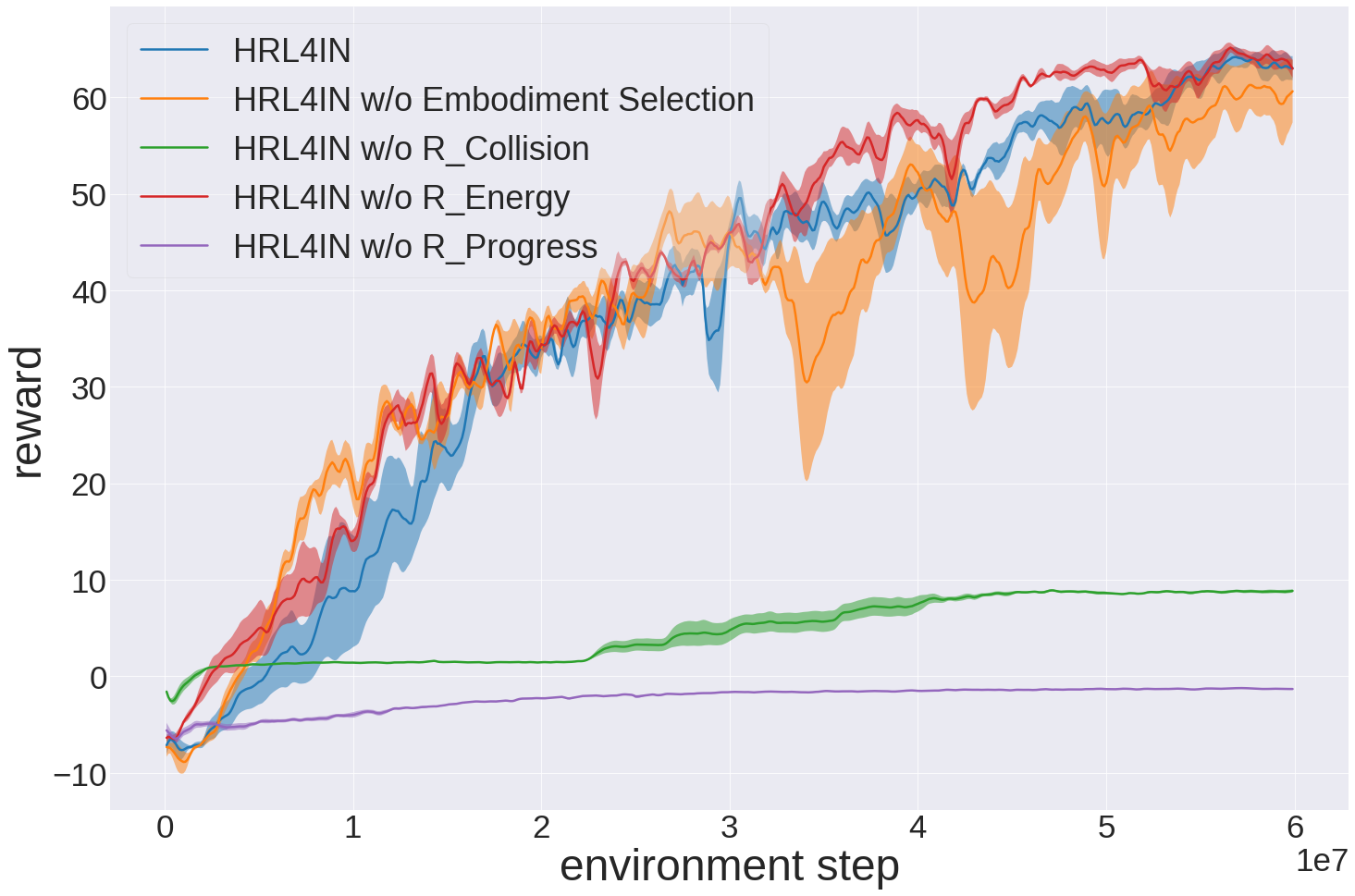}%
\label{sf:gibson_succ}
\caption{}
\end{subfigure}%
~\begin{subfigure}[b]{0.33\textwidth}
\includegraphics[width=0.99\textwidth]{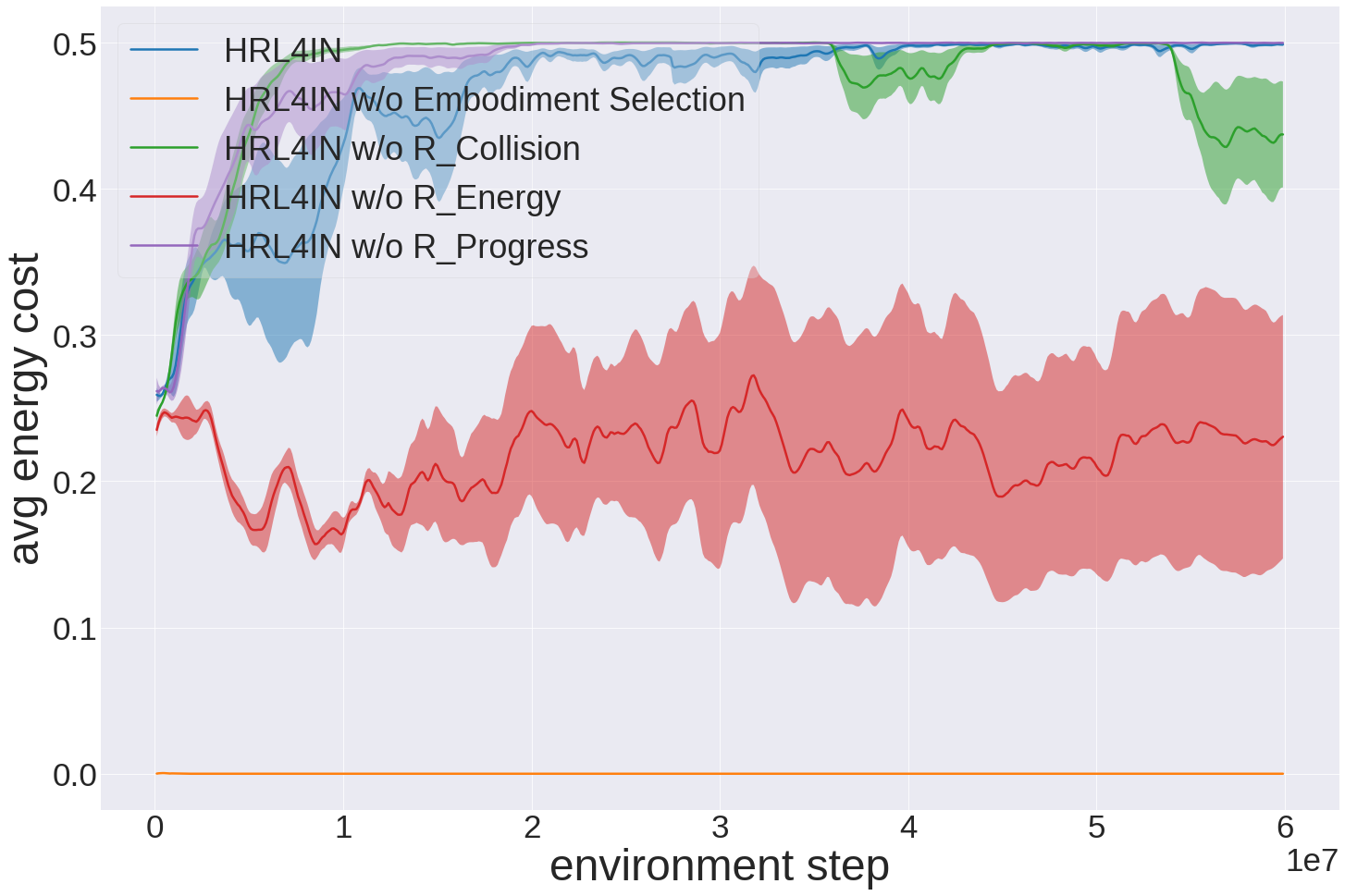}%
\label{sf:gibson_succ}
\caption{}
\end{subfigure}%
\caption{Success rate, reward and energy-saving over time for HRL4IN and its ablated versions in \texttt{Interactive Gibson Environment}. $r_{\mathit{progress}}$ and $r_{\mathit{collision}}$ are essential for task performance; $r_{\mathit{energy}}$ and embodiment selection are essential for energy efficiency.}
\label{fig:abla}
\end{figure}

Fig.~\ref{fig:abla} depicts the training curves of our ablation studies, which complements Table~\ref{t:ablation_studies}. We observe that $r_{\mathit{progress}}$ and $r_{\mathit{collision}}$ are required for the agent to solve the task, and $r_{\mathit{energy}}$ and embodiment selection are required for the agent to learn embodiment selection and solve the task efficiently.

\begin{figure}
\centering
\begin{subfigure}[b]{0.32\textwidth}
\includegraphics[height=3.5cm]{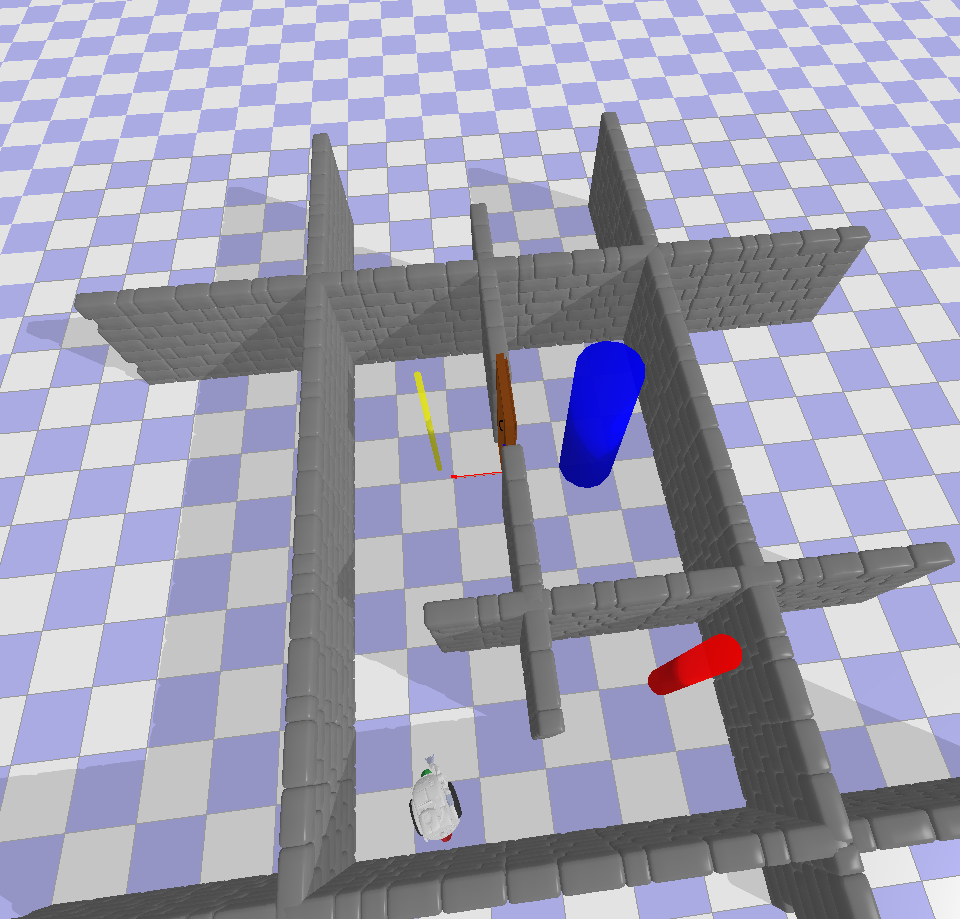}%
\caption{}
\label{sf:sg1}
\end{subfigure}%
~\begin{subfigure}[b]{0.32\textwidth}
\includegraphics[height=3.5cm]{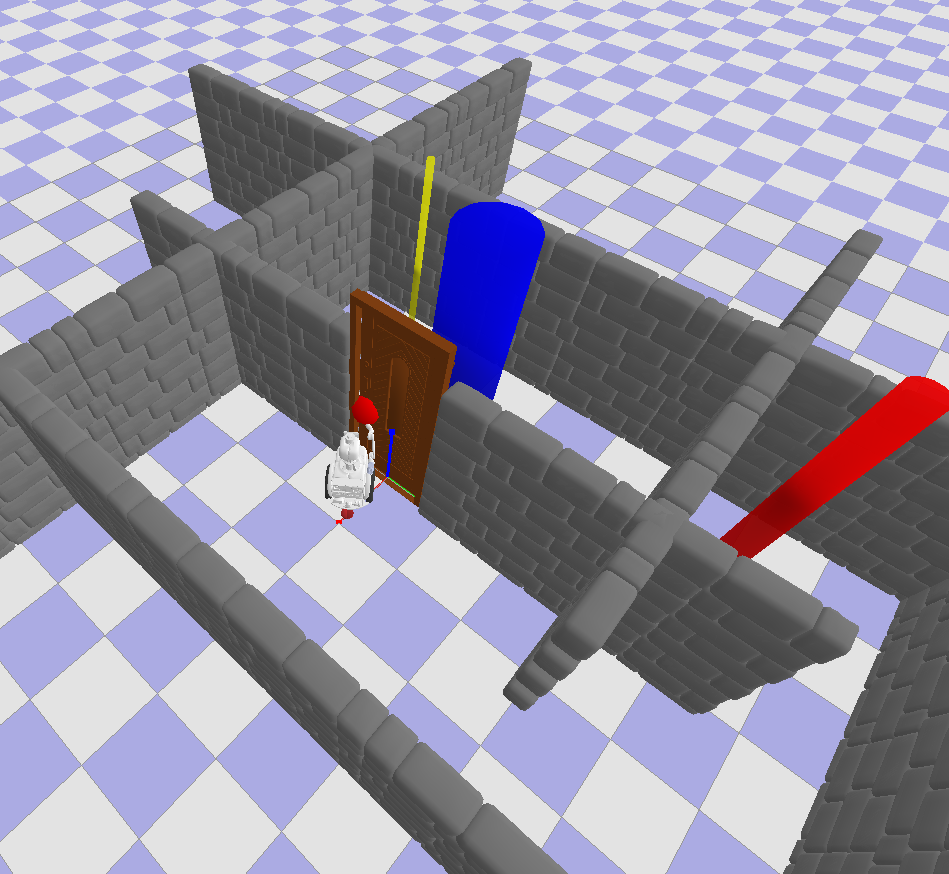}%
\caption{}
\label{sf:sg2}
\end{subfigure}%
~\begin{subfigure}[b]{0.32\textwidth}
\includegraphics[height=3.5cm]{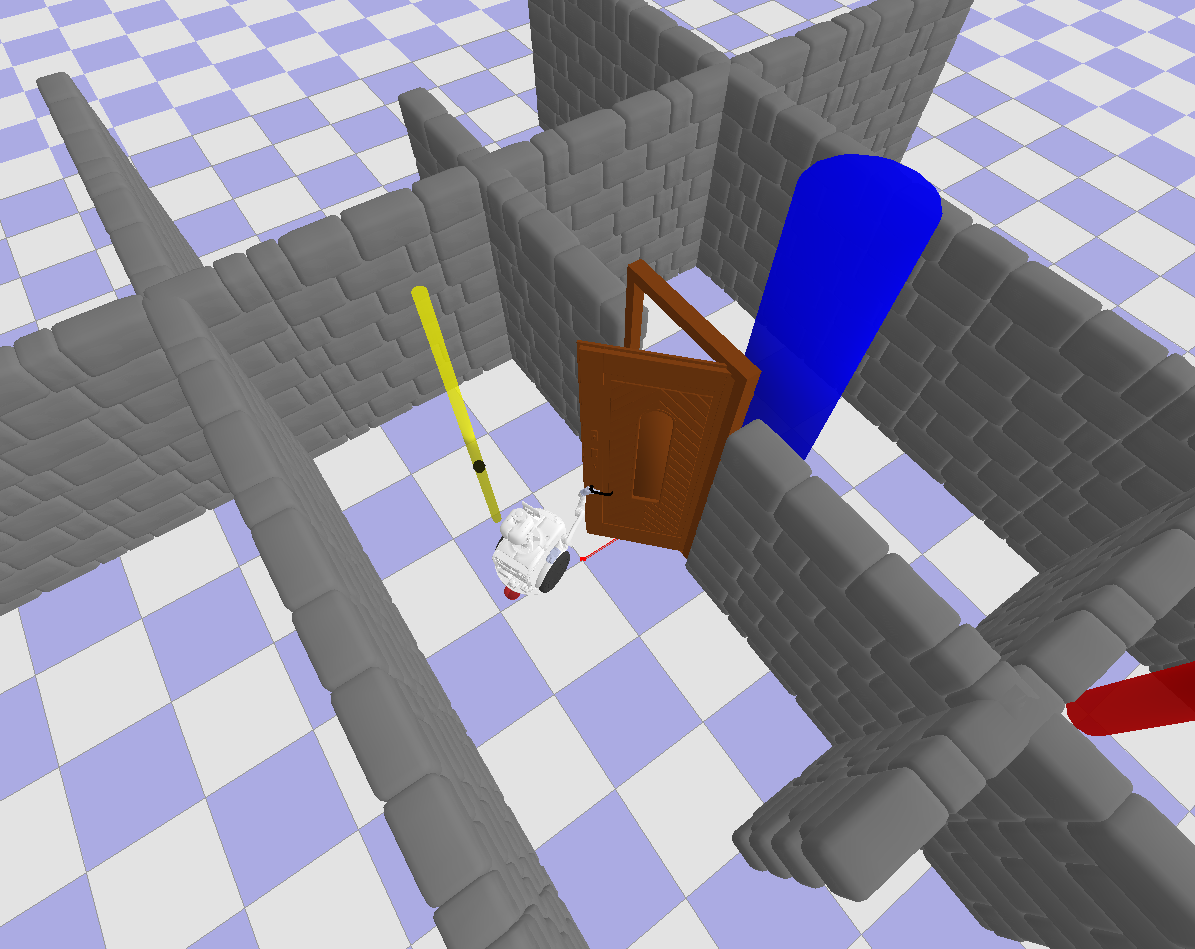}%
\label{sf:sg3}
\caption{}
\end{subfigure}%
\caption{More examples of subgoals set by HRL4IN in \texttt{Interactive Gibson Environment}: (a) a pure navigation (base-only) subgoal is set to guide the robot to move closer to the door; (b) a base+arm subgoal is set to guide the robot to touch the door handle; Although the subgoal is not set directly on the door handle, when the low-level policy tries to achieve it by stretching its arm, the end-effector grabs the door handle; (c) a  base+arm subgoal is set behind the robot so that it can pull the door handle back and open the door.}
\label{fig:add_vis}
\end{figure}

Fig.~\ref{fig:add_vis} shows additional visualizations of subgoals set by the high-level policy of HRL4IN in \texttt{Interactive Gibson Environment}. These visualizations illustrate that the high level policy of HRL4IN learns to command meaningful subgoals to guide the low-level policy to achieve different types of subtasks, including pure navigation, reaching for the door handle and pulling the door open.

\end{appendices}

\end{document}